\documentclass[preprint,12pt]{elsarticle}
\usepackage{graphicx}
\usepackage{amsmath}
\usepackage{amsfonts}
\usepackage[utf8]{inputenc}
\usepackage{soul,color}
\usepackage{hyperref}
\usepackage{multirow}
\usepackage{float}
\usepackage{caption}
\usepackage{subcaption}
\journal{arXiv}
\begin{document}
%%%%%%%%%%%%%%%%%%%%%%%%%%%%%%%%%%%%%%%%%%%%%%%%%%%%%%%%%%%%%%%%%%%%%%%%%%%%%%%%%%%%%%%%%%%%%%%%%%%%%%%%%%
%%%%%%%%%%%%%%%%%%%%%%%%%%%%%%%%%%%%%%%%%%%%%%%%%%%%%%%%%%%%%%%%%%%%%%%%%%%%%%%%%%%%%%%%%%%%%%%%%%%%%%%%%%
\begin{frontmatter}
\title{Determining input variable ranges in Industry 4.0: A heuristic for estimating the domain of a real-valued function or trained regression model given an output range}

\author[add1]{Noelia Oses\corref{cor1}}
\ead{noses@vicomtech.org}
\author[add1]{Aritz Legarretaetxebarria}
\ead{alegarreta@vicomtech.org}
\author[add1]{Marco Quartulli}
\ead{mquartulli@vicomtech.org}
\author[add1]{Igor Garc\'ia}
\ead{iolaizola@vicomtech.org}
\author[add2]{Mikel Serrano}
\ead{mikel.serrano@dominion.es}

\cortext[cor1]{Please address correspondence to Noelia Oses}
\address[add1]{Vicomtech, Paseo Mikeletegi 57, Parque Cient\'ifico y Tecnol\'ogico de Gipuzkoa, 20009 Donostia/San Sebasti\'an, Spain\\ Tel: +[34] 943 30 92 30, Fax: +[34] 943 30 93 93}
\address[add2]{Dominion Solutions, Ibañez de Bilbao 28, 48009 Bilbao, Spain\\ Tel: +[34] 944 79 37 87}
%%%%%%%%%%%%%%%%%%%%%%%%%%%%%%%%%%%%%%%%%%%%%%%%%%%%%%%%%%%%%%%%%%%%%%%%%%%%%%%%%%%%%%%%%%%%%%%%%%%%%%%%%%
%%%%%%%%%%%%%%%%%%%%%%%%%%%%%%%%%%%%%%%%%%%%%%%%%%%%%%%%%%%%%%%%%%%%%%%%%%%%%%%%%%%%%%%%%%%%%%%%%%%%%%%%%%
\begin{abstract}
Industrial process control systems try to keep an output variable within a given tolerance around a target value. PID control systems have been widely used in industry to control input variables in order to reach this goal. However, this kind of Transfer Function based approach cannot be extended to complex processes where input data might be non-numeric, high dimensional, sparse, etc. In such cases, there is still a need for determining the subspace of input data that produces an output within a given range. This paper presents a non-stochastic heuristic to determine input values for a mathematical function or trained regression model given an output range. The proposed method creates a synthetic training data set of input combinations with a class label that indicates whether the output is within the given target range or not. Then, a decision tree classifier is used to determine the subspace of input data of interest. This method is more general than a traditional controller as the target range for the output does not have to be centered around a reference value and it can be applied given a regression model of the output variable, which may have categorical variables as inputs and may be high dimensional, sparse... The proposed heuristic is validated with a proof of concept on a real use case where the quality of a lamination factory is established to identify the suitable subspace of production variable values.
\end{abstract}


\end{frontmatter}
%%%%%%%%%%%%%%%%%%%%%%%%%%%%%%%%%%%%%%%%%%%%%%%%%%%%%%%%%%%%%%%%%%%%%%%%%%%%%%%%%%%%%%%%%%%%%%%%%%%%%%%%%%
%%%%%%%%%%%%%%%%%%%%%%%%%%%%%%%%%%%%%%%%%%%%%%%%%%%%%%%%%%%%%%%%%%%%%%%%%%%%%%%%%%%%%%%%%%%%%%%%%%%%%%%%%%
\section{Introduction}
\label{sec:intro}

Industrial process control systems try to keep an output variable within a given tolerance around a target value. PID control systems have been widely used in industry to control input variables in order to reach this goal. However, this kind of Transfer Function based approach cannot be extended to complex processes where input data might be non-numeric, high dimensional, sparse, etc. In such cases, there is still a need for determining the subspace of input data that produces an output within a given range. Furthermore, in other industrial processes, an output variable might need to be above or below a threshold. Take as an example the case of the eccentricity problem in the seamless steel tube manufacturing industry. An eccentric tube is one for which the circle defined by the outside diameter and the circle defined by the inside diameter of the tube are not concentric, i.e. they have different center points. In seamless steel tube manufacturing, wall-thickness eccentricity is a major problem \cite{Guo2015} and it is common to establish a threshold for the eccentricity of the tubes. In this context, and having a model for the eccentricity, the objective is to determine the values of the input parameters that will produce tubes with eccentricity below the given threshold. This is not an optimization problem as it is not possible to minimize the eccentricity, it is unachievable to manufacture tubes with no eccentricity \cite{Kara2010}, and, even if it were possible, it might not be cost-effective to aim at no eccentricity at all. This paper describes this problem in general terms so that the proposed solution is applicable in a wide range of fields. 

In general, industrial processes often need an output variable to be within a specified range (e.g. within tolerance of a target value or below/above a certain threshold). However, the value of that variable may depend on the values of different input variables and, possibly, on their interactions, too. This can be understood as a generalisation of an open-loop control system where the reference or set point is a range. The relationship between the dependent (output) and the independent (input) variables can be modelled using regression \cite{fisher1922,Aldrich2005} or expressed with a mathematical function. In mathematics, this problem could be stated as follows: given a function $f$ and a set $I$ which is an image of that function, find the set $D$ which is the domain of the function. If the $f$ function has an inverse, then $D$ is the image of the domain $I$ for the inverse function, i.e. $f^{-1}: I \mapsto D$ or $ D = f^{-1}( I )$. However, this solution is contingent on the existence of the inverse function. This is often not the case. Moreover, working with data in Industrie 4.0 \cite{Hermann2015} and the Industrial Internet \cite{Evans2012}, the analytical expression of the $f$ function is, most likely, not known and needs to be inferred from the data (e.g. with an artificial neural network). Therefore, it is necessary to develop a method to obtain $D$ that applies whether the relationship between the inputs and the output is modelled with a mathematical function (invertible or not) or inferred with regression techniques.

This paper proposes a non-stochastic heuristic to determine input values for a mathematical function or a trained regression model given an output range. For this general problem to be described as a control problem, it is sufficient to substitute the target output range by a unique value, i.e. the reference, and a tolerance. The proposed solution can be trivially adjusted to cover that case. Additionally, this method is more general than a traditional controller as the target range for the output does not have to be centered around a reference value and it can be applied given a regression model of the output variable, which may have categorical variables as inputs and may be high dimensional, sparse... In fact, the target output range does not need to be an interval, it can be any subset of the real numbers, a subset that may be connected (e.g. [0,1]) or not (e.g. $[3,3.5] \cup [5,6)$), bounded or not.

The proposed method might have applications in different fields. In Engineering, the method can be applied in tolerance analysis and allocation \cite{Chase1988}. A measured value must be within permissible limits. This measured value is most likely dependent on other engineering parameters. This relationship might be established with regression techniques and, then, the proposed method used to infer the input ranges that will result in the measured value being within the permitted limits.

This paper is structured as follows. Section \ref{sec:relatedwork} presents some related work. Section \ref{sec:problemstatement} provides a formal problem statement. Section \ref{sec:proposedmethod} describes the proposed heuristic in detail. Section \ref{sec:example} uses a simple example to illustrate the application of the proposed method. Section \ref{sec:methodanalysis} analyses the sensibility of the performance of the heuristic to the value of the granularity parameter and to the presence of noise. Section \ref{sec:results} describes a proof-of-concept validation with the results obtained when applying the proposed method to real data collected in a part manufacturing process. Finally, section \ref{sec:conclusions} offers some conclusions.

%%%%%%%%%%%%%%%%%%%%%%%%%%%%%%%%%%%%%%%%%%%%%%%%%%%%%%%%%%%%%%%%%%%%%%%%%%%%%%%%%%%%%%%%%%%%%%%%%%%%%%%%%%
%%%%%%%%%%%%%%%%%%%%%%%%%%%%%%%%%%%%%%%%%%%%%%%%%%%%%%%%%%%%%%%%%%%%%%%%%%%%%%%%%%%%%%%%%%%%%%%%%%%%%%%%%%
\section{Related work}
\label{sec:relatedwork}

There are several patents in which the solution includes an step of \emph{output-to-input propagation}. The     \emph{requirements-based test generation} patent by \cite{Hickman2010} is one of them. This invention automatically generates test cases for the verification of algorithm implementations against its design specifications. The method relies on the data flow diagram that models a system function. In this case, for each block type, the output-to-input propagation specication describes how to determine what values (or ranges) can appear at the block input ports in order to make a given value appear at the specified output port. The patent does not describe how this is done.

The patent by \cite{Li2006} reports on a method for creating dynamic models of etch processes in semiconductor manufacturing. As stated in the patent \emph{'the dynamic process model is used to determine input values that result in a desired output value'}. The method in this patent uses \emph{'the validated dynamic model to optimize process recipes by adjusting input values until the output values predicted by the model match the desired output values as closely as possible under the maximum and minimum value constraints imposed on the process inputs'}. This search to find input values is trial-and-error. This paper presents a method that automatises that process.

Related to a lesser degree is the work on selecting values of input variables in the analysis of output from a computer code \cite{mckayetal1979, mckayetal2000}. In this case, they treat the input variables as random variables and propose three methods of selecting (i.e. sampling) input variable values to analyse the output.

%%%%%%%%%%%%%%%%%%%%%%%%%%%%%%%%%%%%%%%%%%%%%%%%%%%%%%%%%%%%%%%%%%%%%%%%%%%%%%%%%%%%%%%%%%%%%%%%%%%%%%%%%%
%%%%%%%%%%%%%%%%%%%%%%%%%%%%%%%%%%%%%%%%%%%%%%%%%%%%%%%%%%%%%%%%%%%%%%%%%%%%%%%%%%%%%%%%%%%%%%%%%%%%%%%%%%
\section{Problem statement}
\label{sec:problemstatement}

Let $f$ be a real-valued function of $m$ real variables with domain $D \subseteq \mathbb{R}^m$ and image or range $I \subseteq \mathbb{R}$:

\[
f : D \mapsto I
\]

\noindent Let's denote the target range as $ I_t = [y_l, y_u] \subset I$. Let's define as the \emph{positive domain}, $D^+$, the set of points of the feature space for which the output of the function or the estimate of the trained model is within the target output range. 

\[
D^+ = \{ x \in D: f(x) \in I_t \}
\]

\noindent Similarly, the \emph{negative domain}, $D^-$, is the set of points of the feature space for which the output of the function or the estimate of the trained model is outside of the target output range. 

\[
D^- = \{ x \in D: f(x) \notin I_t \}
\]

\noindent The union of these two sets is equal to the domain of the $f$ function: $ D = D^+ \cup D^-$. As a consequence of these definitions, we have that the $f$ function maps the positive domain $D^+$ onto the target range $I_t$:

\[
f: D^+ \mapsto I_t
\]

Finding the positive domain, $D^+$, is the objective and this paper proposes a heuristic for obtaining $\widetilde{D^+}$, an approximation to the positive domain. Even though this problem statement has assumed that there is a mathematical function $f$, the proposed heuristic will work equally well if the $f$ function is replaced by a trained regression model.

%%%%%%%%%%%%%%%%%%%%%%%%%%%%%%%%%%%%%%%%%%%%%%%%%%%%%%%%%%%%%%%%%%%%%%%%%%%%%%%%%%%%%%%%%%%%%%%%%%%%%%%%%%
%%%%%%%%%%%%%%%%%%%%%%%%%%%%%%%%%%%%%%%%%%%%%%%%%%%%%%%%%%%%%%%%%%%%%%%%%%%%%%%%%%%%%%%%%%%%%%%%%%%%%%%%%%
\section{Proposed method}
\label{sec:proposedmethod}

The method proposed in this paper to solve the stated problem is a non-stochastic heuristic. The problem-related parameters of this method are a trained regression model or a mathematical function that relates inputs to outputs, the initial ranges (i.e. intervals) for the inputs, and the target range for the output. In summary, the method proposes to create a synthetic training data set of input combinations with a class label that indicates whether the output (the prediction generated by the trained model or the output generated by the function, given the inputs) is within the given target range. This, then, becomes a classification problem and the output of a decision tree classifier provides exactly what the user needs: combinations of input ranges labeled depending on whether the output will be inside or outside the given range.

\subsection{Training}

The training consists in three steps: constructing a synthetic training data set, training the decision tree classifier, and extracting the products of input ranges from the trained decision tree classifier.

\subsubsection{Construction of the synthetic training data set}
\label{sec:syntheticset}

In addition to the three problem-related parameters mentioned above (a trained model or function that relates inputs to outputs, the initial ranges for the inputs, and the target range for the output), this method also needs a method-related parameter that determines the granularity of the synthetic training set. The granularity parameter will be denoted by $\delta$.

The synthetic training data set will have two parts, the synthetic feature values and the label. The synthetic feature value set is a grid created from the initial ranges for the inputs and the specified granularity ($\delta: \delta>0$). If there are $m$ independent variables, each $X_i$ variable with an initial range $\left[a_i, b_i\right]$ ($ i = 1, \dots, m$), then $V_i$ is the set of synthetic values for $X_i$: 

\[
V_i = \{a_i\} \cup \{b_i\} \cup \{ a_i + n \delta: a_i + n \delta \leq b_i, n \in \mathbb{N} \}
\]

Then, the synthetic feature value set $\mathcal{V}$ is the cartesian product of these one-dimensional synthetic feature value sets:

\[
\mathcal{V} = V_1 \times V_2 \times \dots \times V_m
\]

This paper will sometimes refer to the synthetic feature value set $\mathcal{V}$ as the \emph{grid}.
The class label for each point in the grid depends on the estimate obtained from the trained model or the output of the function applied to the grid point. The class label for the point will be "Inside" if the estimate or output falls within the given target range. Otherwise, the label will be "Outside". Thus, the grid together with the class labels form a labeled data set in a classification problem. The synthetic training data set (or labeled grid) will be denoted by $\mathcal{D}$:

\[
\mathcal{D} = \{ \left(v, l(v) \right): v \in \mathcal{V} \}
\]

\noindent where $l(v)$ denotes the label for $v$. If the target output range is $[y_l, y_u]$ and the function or trained model is denoted by $f$, then the label for a synthetic feature value combination $v$ is determined as follows:

\[
l(v) = 
\begin{cases}
    \text{"Inside"}, & \text{if } f(v) \in  [y_l, y_u] \\
    \text{"Outside"},& \text{otherwise}
\end{cases}
\]

\subsubsection{Train the decision tree classifier}

The decision tree classifier \cite{Breiman1984,quinlan1986} is trained with the synthetic training data set $\mathcal{D}$. We assume that the tree is fully grown, with no limitations on the size of the tree nor constraints on the splitting of the nodes, and that it has not been pruned.

\subsubsection{Approximate positive domain}

The trained tree can be transformed into a set of input range combinations. Each path from the root node to a leaf node determines a product of input ranges. The product of input ranges determined by the $i$-th path will be denoted by $\widetilde{D_i^+}$: 

\[
\widetilde{D_i^+} = [ x_{1i}^l, x_{1i}^u ] \times \dots \times [ x_{mi}^l, x_{mi}^u ]
\]

\noindent where $x_{ji}^l$ and $x_{ji}^u$ are the lower and upper endpoints of the interval for the $j$-th variable for the $i$-th path, respectively. Note that, for ease of reading, all the intervals in $\widetilde{D_i^+}$ have been written as closed intervals, but some or all could be open or mixed (left/right-closed, left/right-open) intervals.

If the tree has not been pruned or constrained in any way, then each path leads to a unique class, i.e. either all of the model estimates or function outputs for all of the synthetic training data points in the product of input ranges are within the target output range or they are not. This allows selecting those input range combinations for which the estimate of the model is within the given output range. $\widetilde{D^+}$, the approximate positive  domain (APD), is the union of all the input range products for which the output is within the target output range.

\[
\widetilde{D^+} = \bigcup_{i: \text{the leaf node of the i-th path has class label "Inside"}} \widetilde{D_i^+}
\]

\subsection{Testing}

At this point, the trained tree will have 100\% accuracy on the synthetic training data set. However, the extracted input ranges depend on the granularity parameter of the method. In addition, if a trained model was used to produce the estimate for each grid point, that estimate will also have some error. Therefore, the accuracy of the calculated input ranges must be tested on a test data set. This might possibly be the same test set used to evaluate the fit of the trained model, if a regression model was used, or a synthetic test set generated by uniformly sampling from the given initial input ranges. 

The test set is used to construct a contingency table very similar to a confusion matrix. For each point in the test set, the feature values are checked to see if they fall within one of the input range combinations extracted from the trained tree. Similarly, each observed output is checked against the target output range. After these two checks, the appropriate count of the contingency table (table \ref{tab:evaluationexample}) is updated.

\begin{table}[H]
\centering
\caption{Evaluation of the extracted input ranges on a test set}
\label{tab:evaluationexample}
\begin{tabular}{c|c|c|c|}
\multicolumn{1}{c}{} & \multicolumn{1}{c}{} & \multicolumn{2}{c}{OUTPUT} \\ 
\cline{3-4}
\multicolumn{1}{c}{} & & Inside & Outside  \\
\cline{2-4}
\multirow{2}{*}{INPUT} & Inside & True Positive & False Positive \\
\cline{2-4}
& Outside & False Negative & True Negative \\
\cline{2-4}
\end{tabular}
\end{table}

The first row of table \ref{tab:evaluationexample} has a greater importance than the second, as once the input ranges are determined the input combinations provided to the process will, presumably, always be within these ranges. Therefore, a cross-validation approach \cite{Kohavi1995} can be used in order to select a granularity parameter for which the proportion of outputs outside the target range given inputs within the extracted ranges is below some predetermined threshold. In this paper, the true positive rate (eq. \ref{eq:trueposrate}) of the pseudo-confusion matrix shown in table \ref{tab:evaluationexample} will be used as the evaluation metric.

\begin{equation}
\text{True Positive Rate} = \frac{\text{True Positive}}{\text{True Positive} + \text{False Positive}}
\label{eq:trueposrate}
\end{equation}

\subsection{Improving the true positive rate}
\label{sec:improvingtpr}

The true positive rate (TPR) can be improved by taking only those input range products for which the leaf node has class label "Inside" and which are contained in the positive domain, i.e. $\widetilde{D_i^+} \subset D^+$. Let's denote this subset of $\widetilde{D^+}$ by $\widehat{D^+} = \bigcup_{i: \widetilde{D_i^+} \subset D^+ \text{and the leaf node of the i-th path has class label "Inside"}} \widetilde{D_i^+}$. If $\widehat{D^+}$ is not the empty set, then its TPR is 1. Note that $\widehat{D^+}$ will not cover the positive domain. If providing a good cover of the positive domain is important, it will be better to improve the TPR by using a finer synthetic training data set (i.e. using a smaller value of the granularity parameter $\delta$). 

In practice, the problem will be determining whether for a given $i$ $\widetilde{D_i^+}$ is a subset of the positive domain (i.e. $\widetilde{D_i^+} \subset D^+ $). One way to test this condition is to create a synthetic data set within $\widetilde{D_i^+}$, as described in section \ref{sec:syntheticset}, and calculate its TPR. If the TPR is less than 1, then $\widetilde{D_i^+}$ is not a subset of the positive domain. However, it is possible to obtain a TPR of 1 even if $\widetilde{D_i^+}$ is not a subset of the positive domain if the granularity parameter chosen is not sufficiently small.

% \subsection{Reducing the number of input range combinations}
% 
% In most cases, the method so far will lead to a lengthy list of input range products. One possible way to simplify this result is by pruning \cite{Breiman1984,Quinlan1987,Bohanec1994} the tree. The pruning will most likely have an impact on the accuracy of the extracted input ranges and, thus, again a cross-validation approach might be necessary to choose the level of pruning. Often, the pruning method discussed is known as 'post-pruning', as it allows the tree to grow until it perfectly fits the training data and, then, the tree is pruned. Other methods for simplifying the tree might be setting the minimum number of samples required at a leaf node or setting the maximum depth of the tree. These other methods are known as 'pre-pruning'; they stop the tree from growing before it fits the training data perfectly. The latter approaches are usually less accurate because it is not trivial to determine when to stop growing the tree without having any knowledge of the full-size tree.

%%%%%%%%%%%%%%%%%%%%%%%%%%%%%%%%%%%%%%%%%%%%%%%%%%%%%%%%%%%%%%%%%%%%%%%%%%%%%%%%%%%%%%%%%%%%%%%%%%%%%%%%%%
%%%%%%%%%%%%%%%%%%%%%%%%%%%%%%%%%%%%%%%%%%%%%%%%%%%%%%%%%%%%%%%%%%%%%%%%%%%%%%%%%%%%%%%%%%%%%%%%%%%%%%%%%%
\section{Illustrative example}
\label{sec:example}

\begin{figure}[H]
\centering
\includegraphics[width=0.75\textwidth]{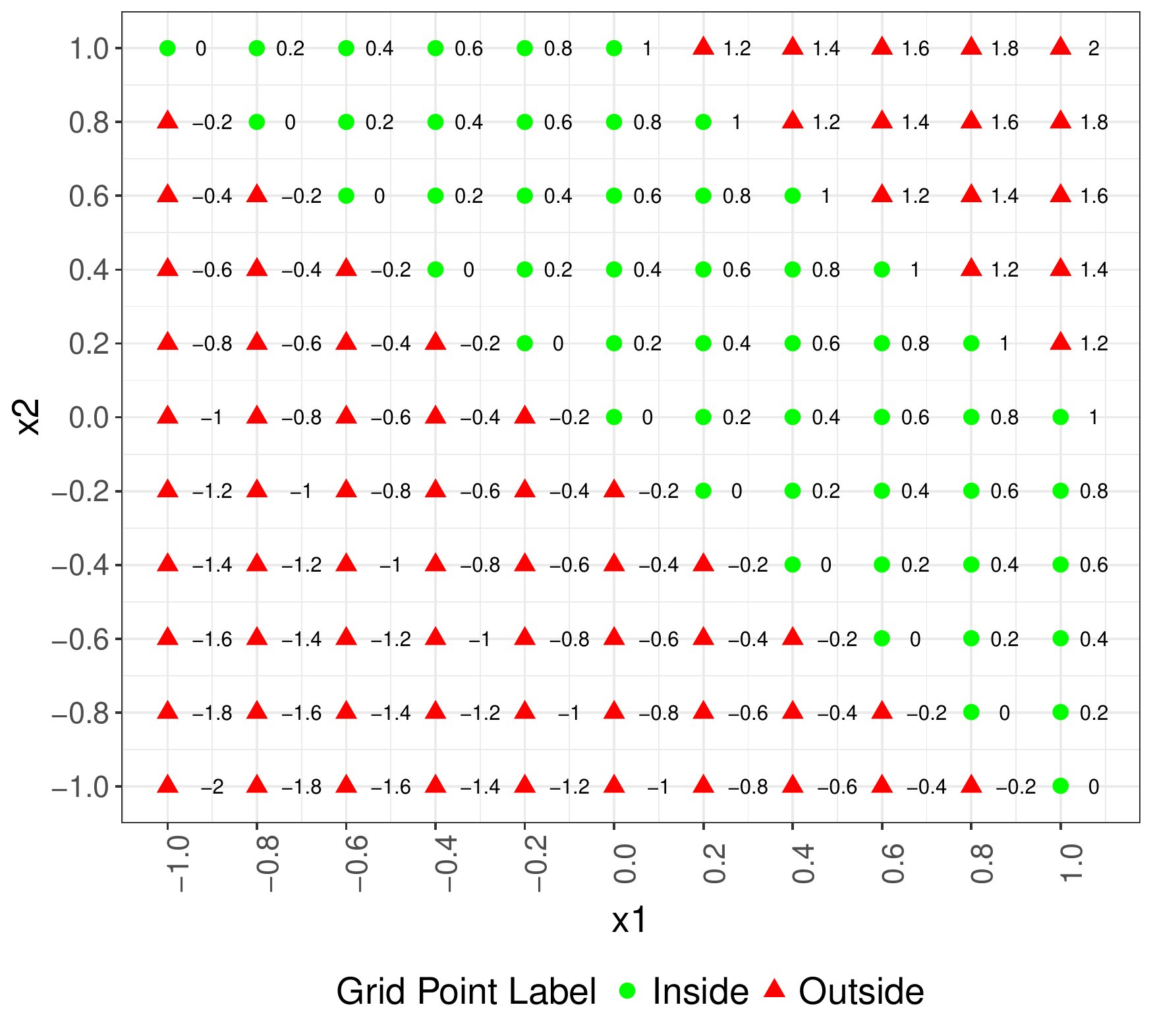}
\caption{Example synthetic training data grid for two independent variables}
\label{fig:examplegrid}
\end{figure}

A simple function will be used for demonstration purposes. The function will be $y = x_1 + x_2$ and the independent variables, $x_1$ and $x_2$, will both have initial range [-1, 1]. The target range for the output variable $y$ will be [0, 1]. Figure \ref{fig:examplegrid} shows the training grid for this example when the granularity parameter is set to 0.2. The numbers to the right of the symbols representing the points in the grid show the value that the function takes at that point. These values are rounded to two decimal places in the plot to make it easier to read.

\begin{figure}[H]
\centering
\includegraphics[width=0.75\textwidth]{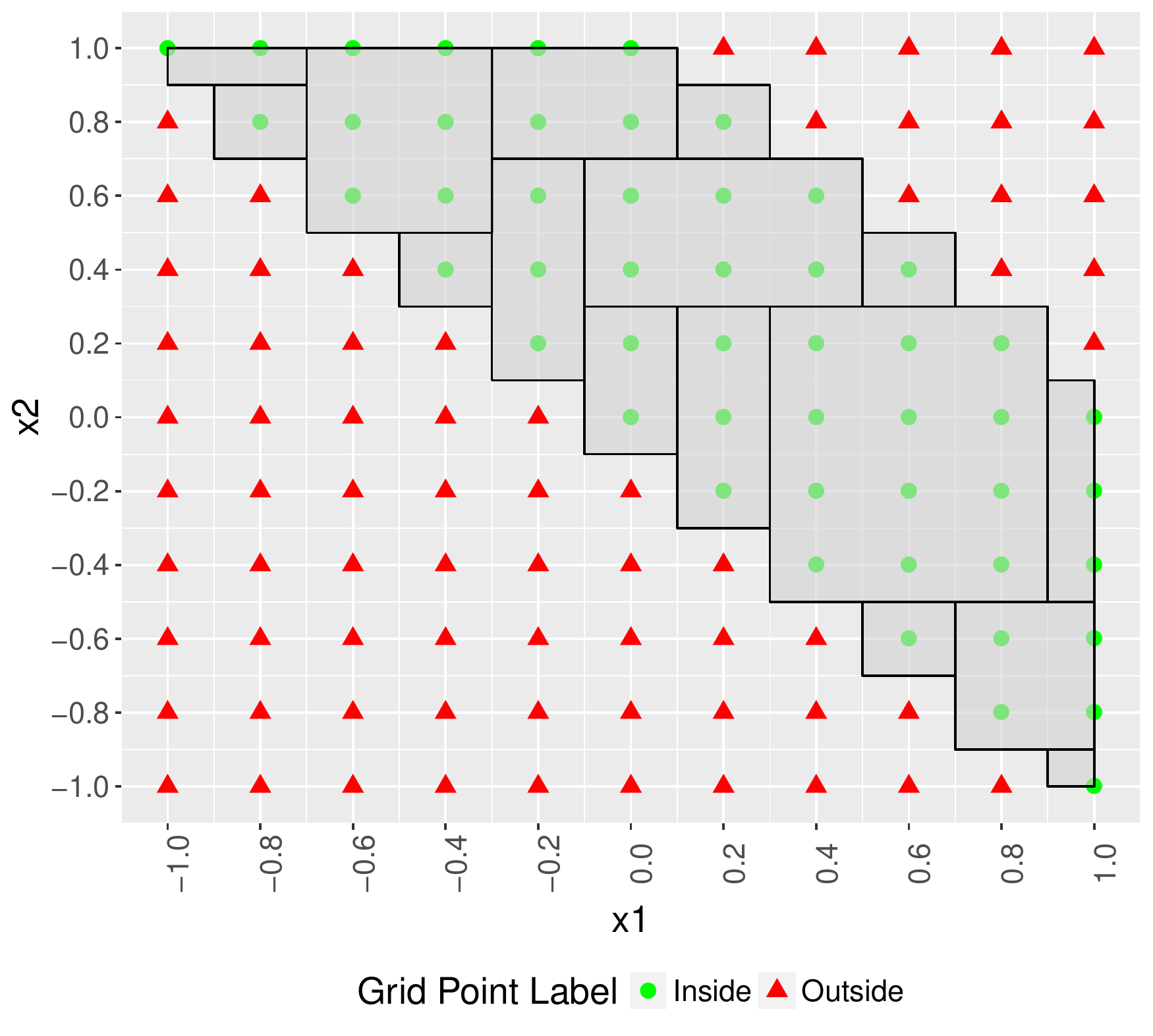}
\caption{Approximate positive domain plotted on top of the same synthetic training data grid used to estimate it}
\label{fig:examplegridextractedintervals}
\end{figure}

Figure \ref{fig:examplegridextractedintervals} shows the approximate positive domain, $\widetilde{D^+}$, represented by the rectangles overlaid on the same synthetic training data set with which they were estimated.

\begin{figure}[H]
\centering
\includegraphics[width=0.75\textwidth]{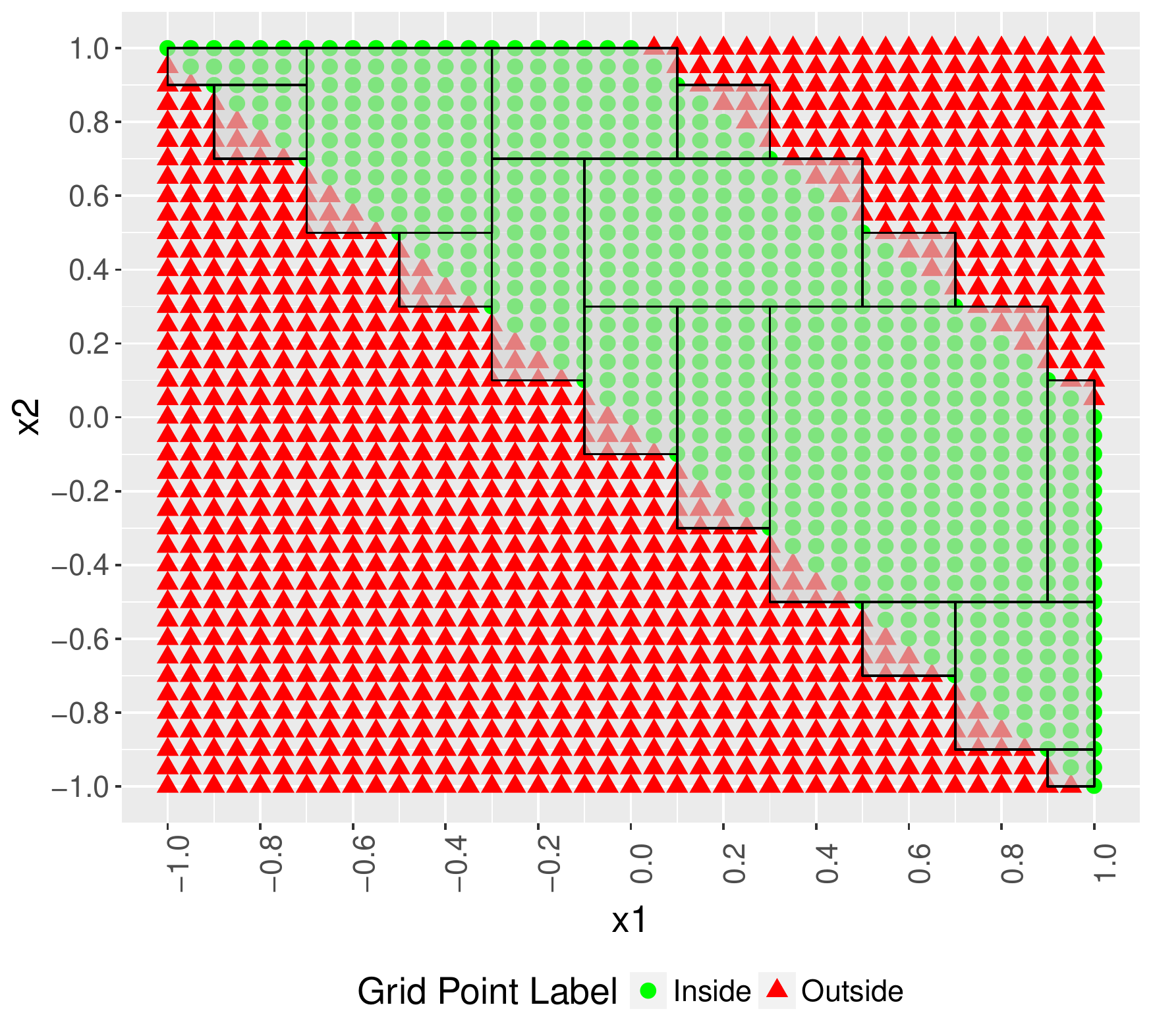}
\caption{Approximate positive domain from fig. \ref{fig:examplegridextractedintervals} plotted on top of a synthetic training data grid created with the granularity parameter set to 0.05}
\label{fig:examplegridextractedintervals_finegrain}
\end{figure}

Figure \ref{fig:examplegridextractedintervals_finegrain} shows the same approximate positive domain as above (fig. \ref{fig:examplegridextractedintervals}) overlayed on top of a synthetic training data set created with the granularity parameter set to 0.05. It can be observed that the approximate positive domain does not fit this synthetic training data set well. In this case, as the function is very simple, it is easy to visualise the positive domain as the region defined by the $x_1+x_2 \geq 0$ and $x_1+x_2 \leq 1$ equations. Therefore, it can be observed that the smaller the granularity parameter value the better the grid represents the positive and negative domains.

In order to evaluate the approximate positive domain numerically, a test set has been created by sampling from the uniform distribution in the [-1, 1] range 10000 times for each independent variable. Table \ref{tab:exampletestset} shows the first five rows of the illustrative test data. 

% latex table generated in R 3.4.4 by xtable 1.8-2 package
% Mon Apr  1 10:06:11 2019
\begin{table}[H]
\centering
\caption{Illustrative test data set for this example (first five rows)} 
\label{tab:exampletestset}
\begingroup\scriptsize
\begin{tabular}{rrr}
  \hline
x1 & x2 & x1+x2 \\ 
  \hline
0.85 & -0.20 & 0.65 \\ 
  0.84 & 0.28 & 1.11 \\ 
  0.33 & 0.83 & 1.16 \\ 
  -0.02 & 0.39 & 0.36 \\ 
  0.48 & 0.72 & 1.20 \\ 
   \hline
\end{tabular}
\endgroup
\end{table}
Table \ref{tab:exampleconfmatrix} shows the corresponding evaluation confusion matrix for the approximate positive domain, which has a true positive rate of 0.843. Even though the function in this example is very simple, the approximate positive domain is not perfect. One factor contributing to this is that the synthetic training set has been constructed using a granularity parameter of 0.2. The impact of the granularity parameter on the results will be studied next.

\begin{table}[H]
\centering
\caption{Evaluation of the approximate positive domain on a test set}
\label{tab:exampleconfmatrix}
\begin{tabular}{c|c|c|c|}
\multicolumn{1}{c}{} & \multicolumn{1}{c}{} & \multicolumn{2}{c}{OUTPUT} \\ 
\cline{3-4}
\multicolumn{1}{c}{} & & Inside & Outside  \\
\cline{2-4}
\multirow{2}{*}{INPUT} & Inside & 3793 & 708 \\
\cline{2-4}
& Outside & 0 & 5499 \\
\cline{2-4}
\end{tabular}
\end{table}

%%%%%%%%%%%%%%%%%%%%%%%%%%%%%%%%%%%%%%%%%%%%%%%%%%%%%%%%%%%%%%%%%%%%%%%%%%%%%%%%%%%%%%%%%%%%%%%%%%%%%%%%%%
%%%%%%%%%%%%%%%%%%%%%%%%%%%%%%%%%%%%%%%%%%%%%%%%%%%%%%%%%%%%%%%%%%%%%%%%%%%%%%%%%%%%%%%%%%%%%%%%%%%%%%%%%%
\section{Sensitivity analysis of the performance of the method}
\label{sec:methodanalysis}

Two experiments have been conducted to study the sensitivity of the proposed method. In the first case, the same experiment is conducted repeatedly for different values of the granularity parameter. In the second case, noise has been added to study its impact on the quality of the results. The experiments perform positive domain approximation and evaluation. The experiments have been carried out with two independent variables so that the results can be shown graphically. The general experiment setup is as follows. Four different functions have been used: $x_1+x_2$, $x_1^2 + x_2^2$, $ \sin (x_1) + \cos (x_2)$, and $ \log (|x_1|+|x_2|)$. The initial range for the independent variables is [-1, 1] for both $x_1$ and $x_2$. Note that the $x_1^2 + x_2^2$ function is not invertible in the given [-1, 1] domain. The target range for the output variable is $I_t = [0, 1]$ at all times in these experiments. Five test sets have been created and the mean TPR is used as the evaluation metric. For each test set and independent variable, 10000 samples have been drawn from the uniform distribution on the variable's initial range of [-1, 1]. 

\subsection{Granularity parameter}

\begin{figure}[H]
\centering
\includegraphics[width=\linewidth]{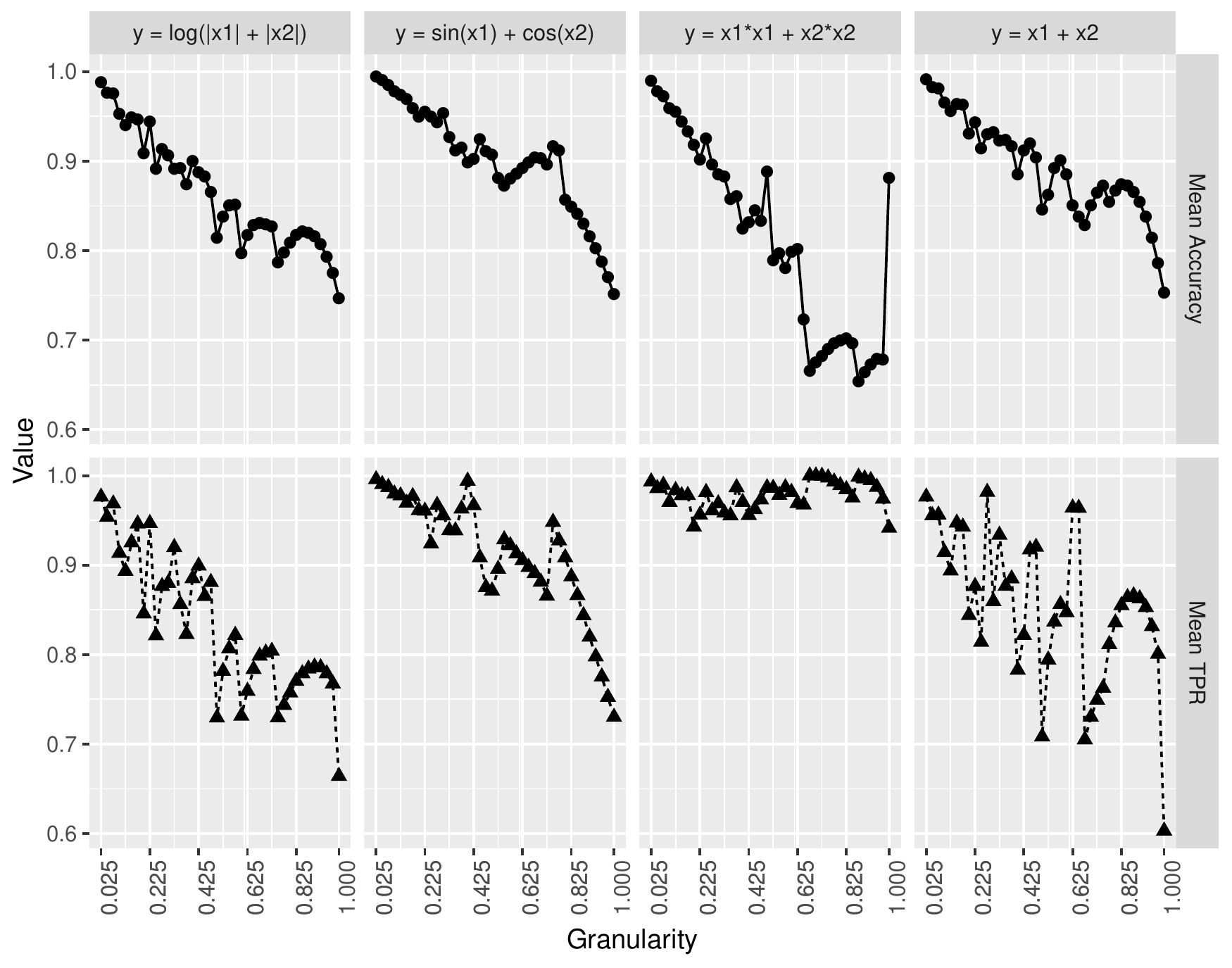}
\caption{Impact of the granularity of the synthetic training data set on the accuracy and true positive rate (TPR) of the approximate positive domains}
\label{fig:granularity}
\end{figure}

An experiment has been run in order to analyse the impact of the granularity parameter on the quality of the results. Figure \ref{fig:granularity} shows the mean accuracy and mean TPR of the approximate positive domain for the different values of the granularity parameter by function. Predictably, the accuracy is highest for the lowest value of the granularity parameter. However, although the mean TPR tends to decrease as the granularity parameter increases, this is not always the case. Additionally, it might be surprising at first sight that the proposed method does not perform best on the simplest function, $y = x_1+x_2$.

As the value of the granularity parameter decreases, the synthetic training data set provides a better approximation of the actual positive and negative sets. Therefore, the smaller the granularity parameter the higher the accuracy achieved. The TPR depends on the value of the granularity parameter but also on the geometry and topology of the positive domain. The geometry and topology of the positive domain depend on all three problem parameters: function, initial input ranges, and target output range. Note that the approximate positive domain, $\widetilde{D^+}$, is always the union of cartesian products of intervals and, thus, it has a particular geometry. 

\begin{figure}[H]
\centering
\begin{subfigure}{.5\textwidth}
  \centering
  \includegraphics[width=\linewidth]{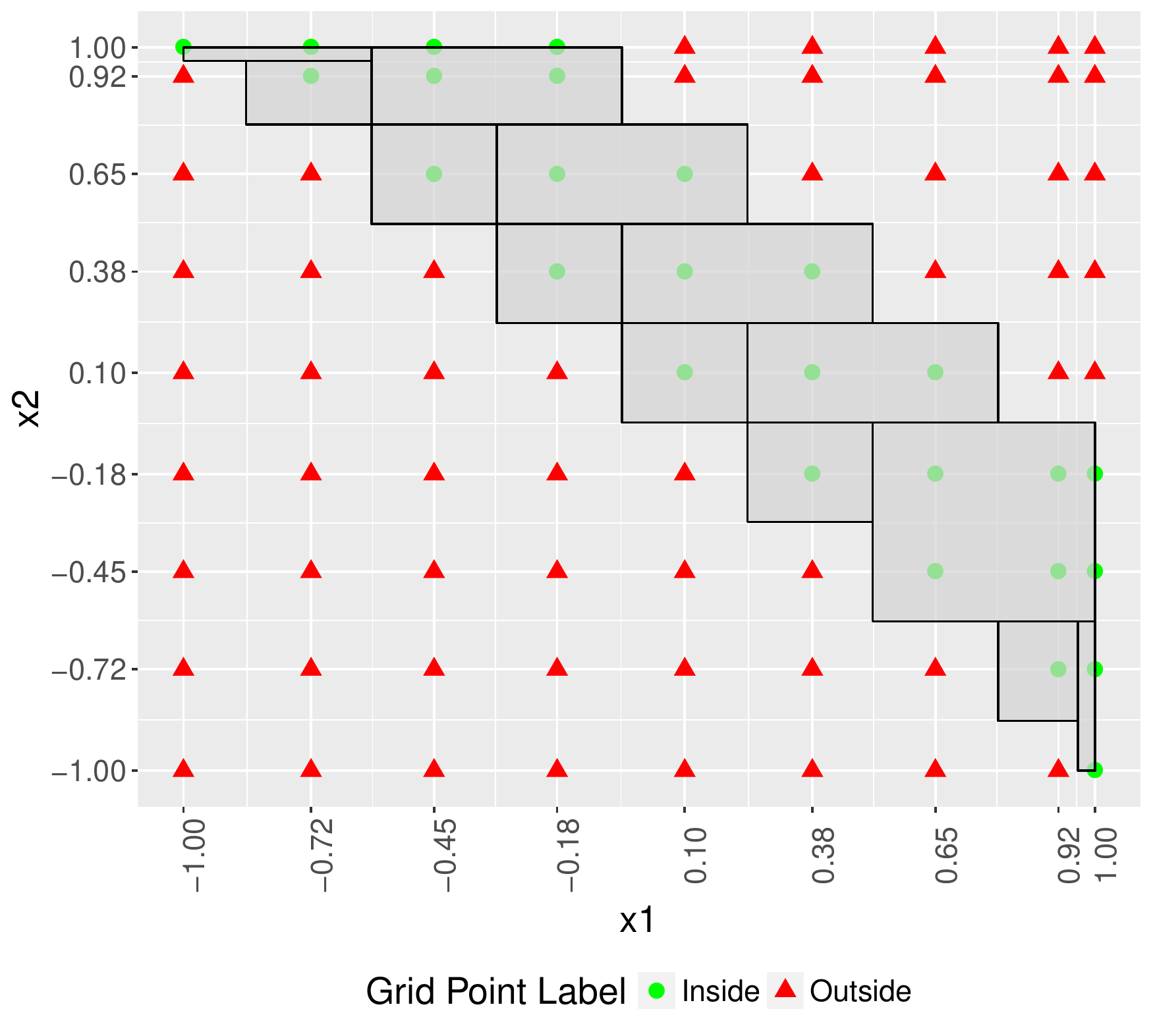}
  \caption{Granularity 0.275}
  \label{fig:linearintervals065grid065}
\end{subfigure}%
\begin{subfigure}{.5\textwidth}
  \centering
  \includegraphics[width=\linewidth]{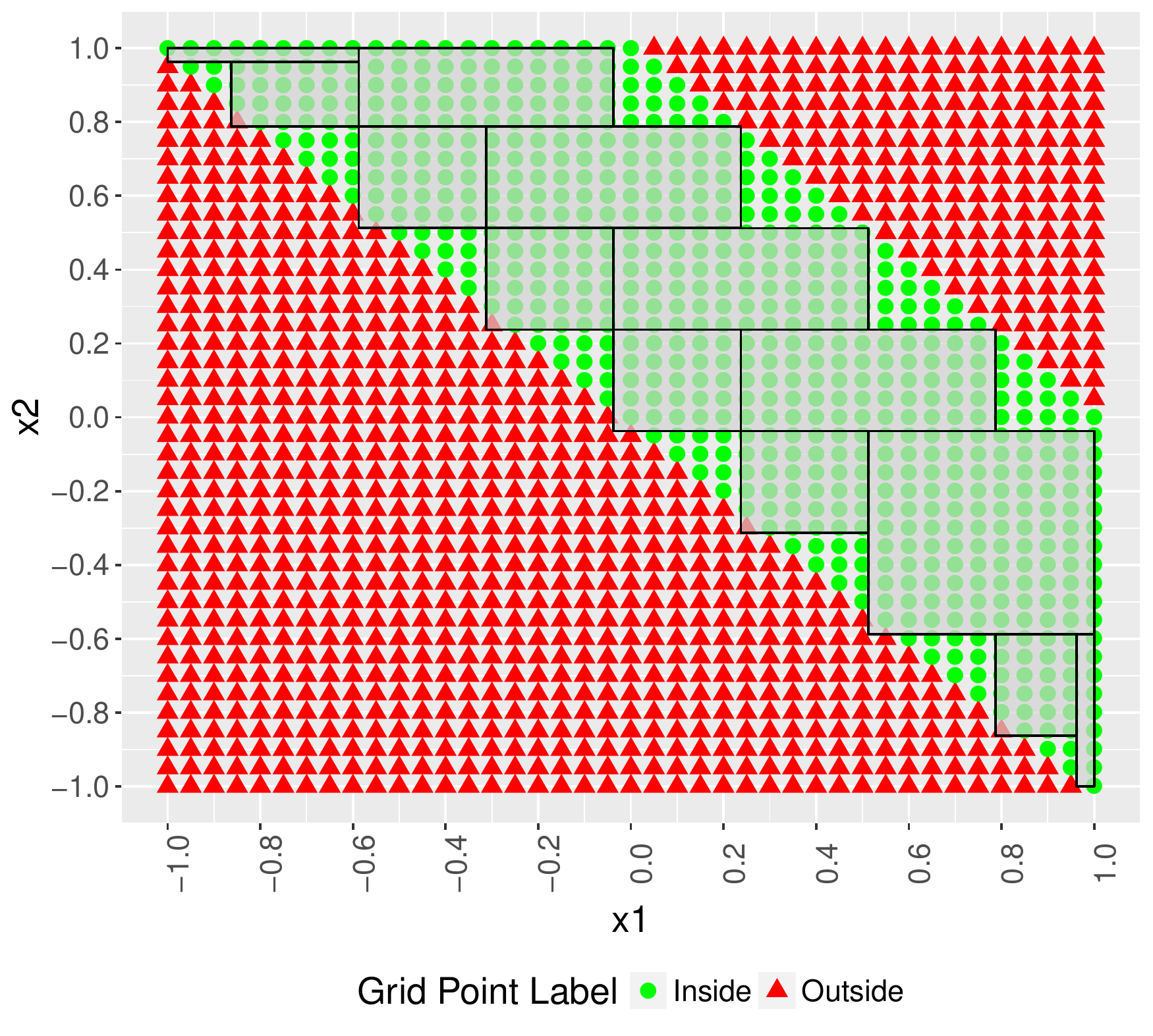}
  \caption{Input ranges' granularity 0.275, grid granularity 0.05}
  \label{fig:linearintervals065grid005}
\end{subfigure}
\begin{subfigure}{.5\textwidth}
  \centering
  \includegraphics[width=\linewidth]{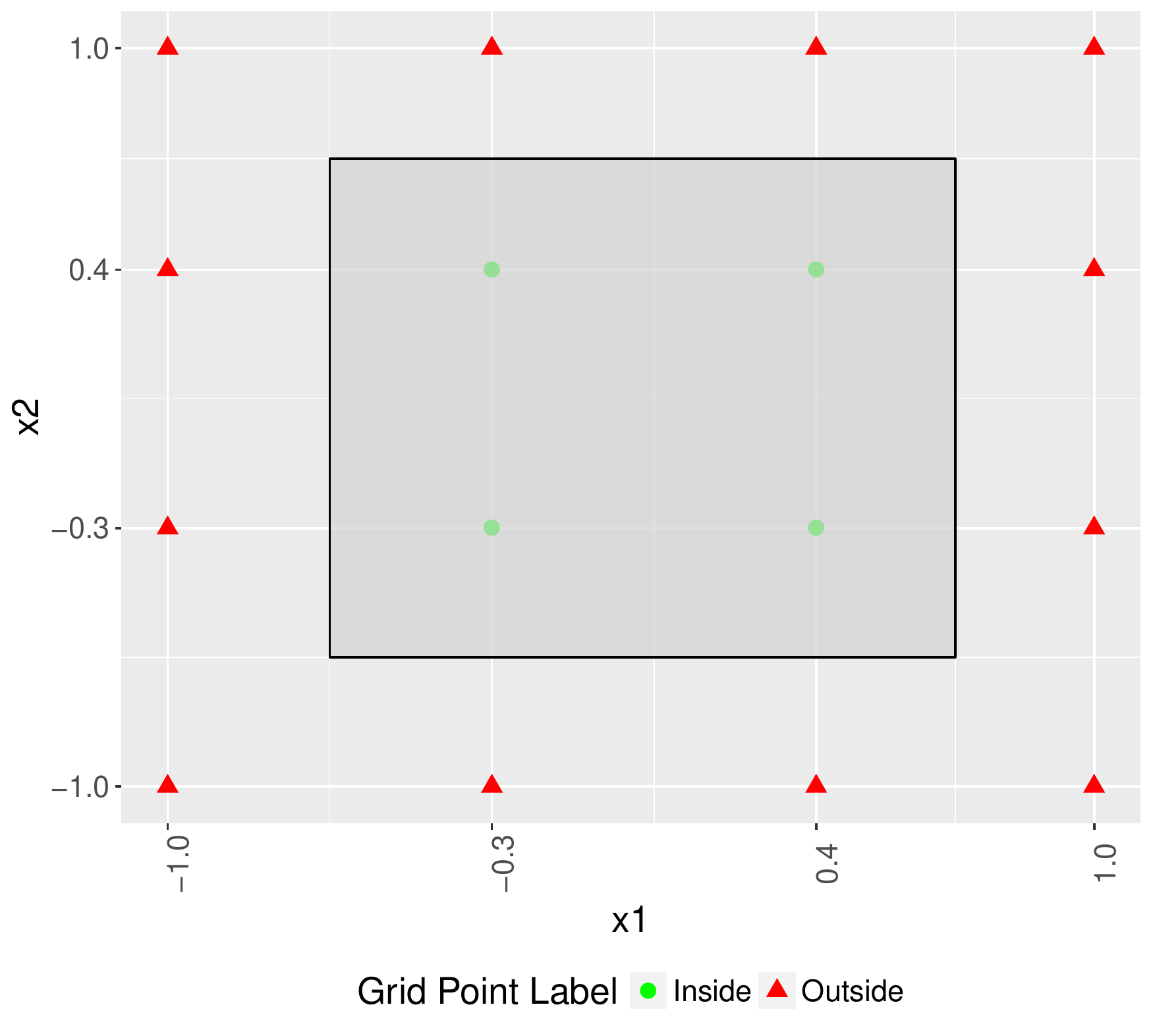}
  \caption{Granularity 0.7}
  \label{fig:sumofsquaresintervals07grid07}
\end{subfigure}%
\begin{subfigure}{.5\textwidth}
  \centering
  \includegraphics[width=\linewidth]{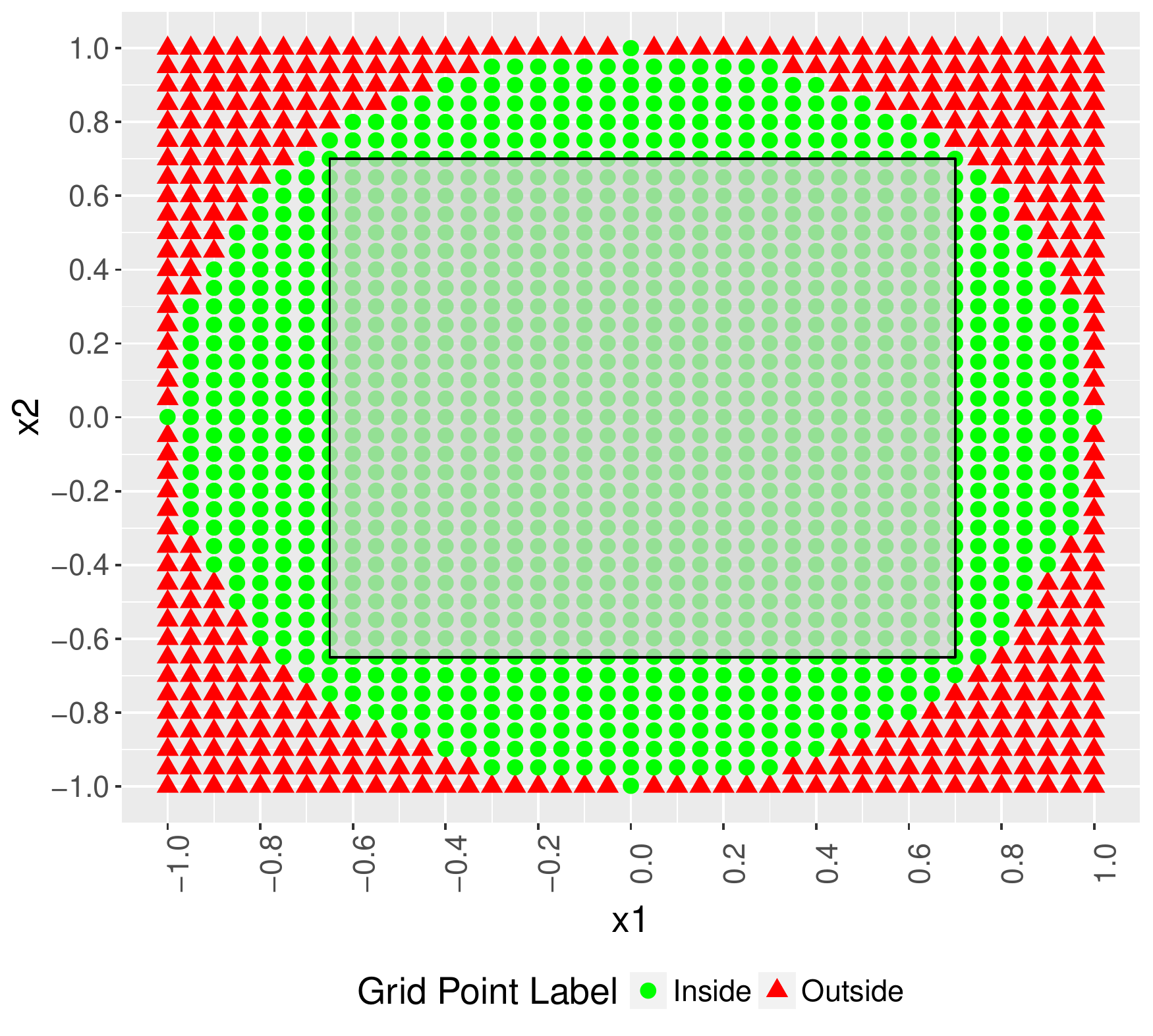}
  \caption{Input ranges' granularity 0.7, grid granularity 0.05}
  \label{fig:sumofsquaresintervals07grid005}
\end{subfigure}
\caption{Approximate positive domains for $y = x_1 + x_2$ (top) and $y = x_1^2 + x_2^2$ (bottom). Left column: plots show the approximate positive domains plotted on top of the synthetic training data grid used for extracting them. Right column: same approximate positive domains as on left column plotted on top of the synthetic training data set for granularity set to 0.05}
\label{fig:extractedintervalsTPRexplanation}
\end{figure}

Figure \ref{fig:extractedintervalsTPRexplanation} illustrates this point with two examples of approximate positive domains, one for $y = x_1 + x_2$ and one for $y = x_1^2 + x_2^2$. The plots on the left column of the figure (\ref{fig:sumofsquaresintervals07grid07} and \ref{fig:linearintervals065grid065}) show the approximate positive domains plotted on top of the synthetic training data grid used for extracting them. These plots help understand how the decision tree classifier arrived at the result. The plots on the right column of the figure (\ref{fig:sumofsquaresintervals07grid005} and \ref{fig:linearintervals065grid005}) show the same approximate positive domains as on the left column but plotted on top of the synthetic training data set for granularity set to 0.05. The latter grid provides a better approximation of the actual positive and negative sets and, thus, it provides a graphical clue as to what the accuracy and TPR of the approximate positive domain will be. It can be observed from figure \ref{fig:sumofsquaresintervals07grid005} that the approximate positive domain for function $y = x_1^2 + x_2^2$ and granularity value $\delta = 0.7$ is strictly contained within the positive domain. Therefore, the TPR for this approximate positive domain is 1, although the accuracy will not be good. If the user is only concerned with ensuring that the output value is within the given target range and does not need to have a good approximation to the actual positive domain, then this is an excellent result. Figure \ref{fig:linearintervals065grid005} shows a similar situation with the function $y = x_1 + x_2$ and $\delta = 0.275$. These plots also show why this method achieves a better TPR for $y = x_1^2 + x_2^2$ than for $y = x_1 + x_2$. Even though the latter function is simpler and the former is not invertible in the given initial domain, the geometry of the positive domain is simpler in the former case.

\begin{figure}[H]
\centering
\begin{subfigure}{.5\textwidth}
  \centering
  \includegraphics[width=\linewidth]{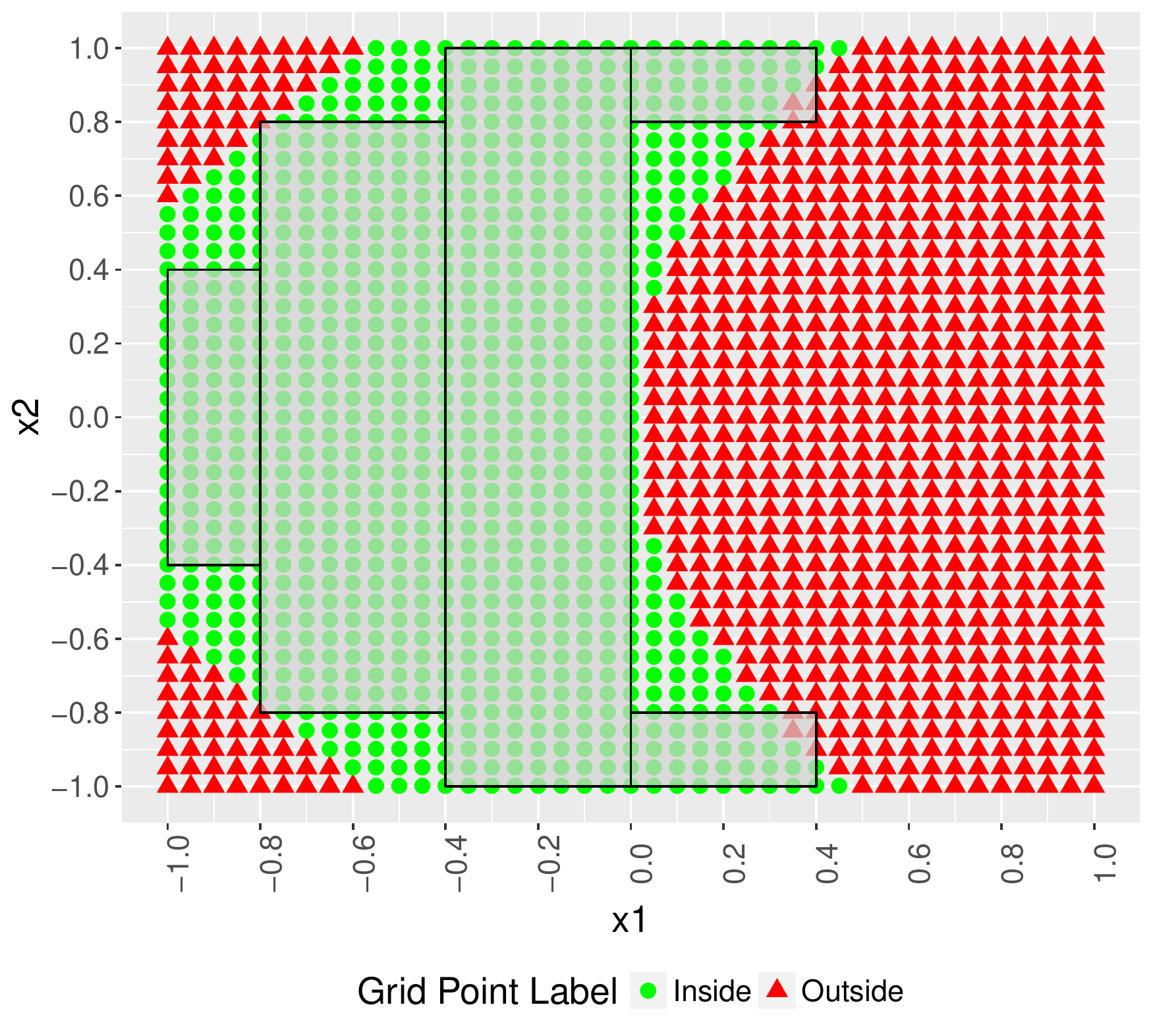}
  \caption{$y = \sin (x_1) + \cos (x_2)$, $\delta = 0.4$}
  \label{fig:sinpluscosintervals04grid005}
\end{subfigure}%
\begin{subfigure}{.5\textwidth}
  \centering
  \includegraphics[width=\linewidth]{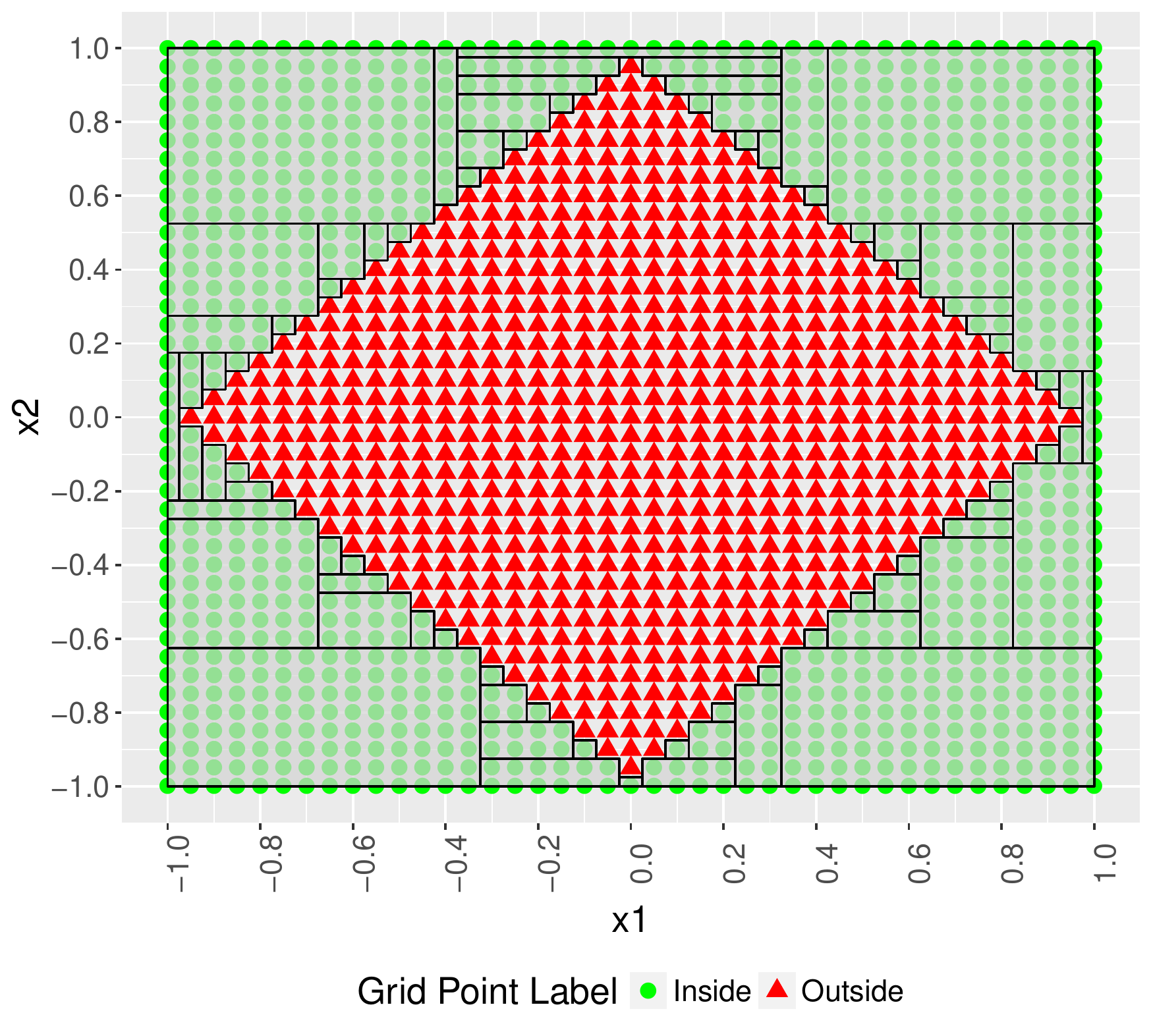}
  \caption{$y = \log (|x_1| + |x_2|)$, $\delta = 0.05$}
  \label{fig:logsumabsdelta005}
\end{subfigure}
\caption{Approximate positive domains plotted on top of the synthetic training data set for granularity set to 0.05}
\label{fig:extractedintervalslasttwo}
\end{figure}

Figure \ref{fig:extractedintervalslasttwo} shows the approximate positive domains with the highest TPR for function $y = \sin (x_1) + \cos (x_2)$ (\ref{fig:sinpluscosintervals04grid005}) and function $y = \log (|x_1| + |x_2|)$ (\ref{fig:logsumabsdelta005}) for illustrative purposes.

\subsection{Noise}

There are two main sources of noise. On the one hand, the values of the independent variables that are specified might not be the values that are actually used in the process. For example, an operator might set a control of an actuator to open a valve to 60\% but the control might have some measurement error and might open the valve in the (60 $\pm$ 1)\% range. On the other hand, the mapping from inputs to outputs might introduce error. For example, a regression model that has been trained with data will have an error associated with the produced estimates. In some cases, both sources of noise might be present. Two experiments have been run in order to analyse the impact of the two sources of noise on the quality of the results. 

\subsubsection{Noisy output} 

\begin{figure}[H]
\centering
\includegraphics[width=\linewidth]{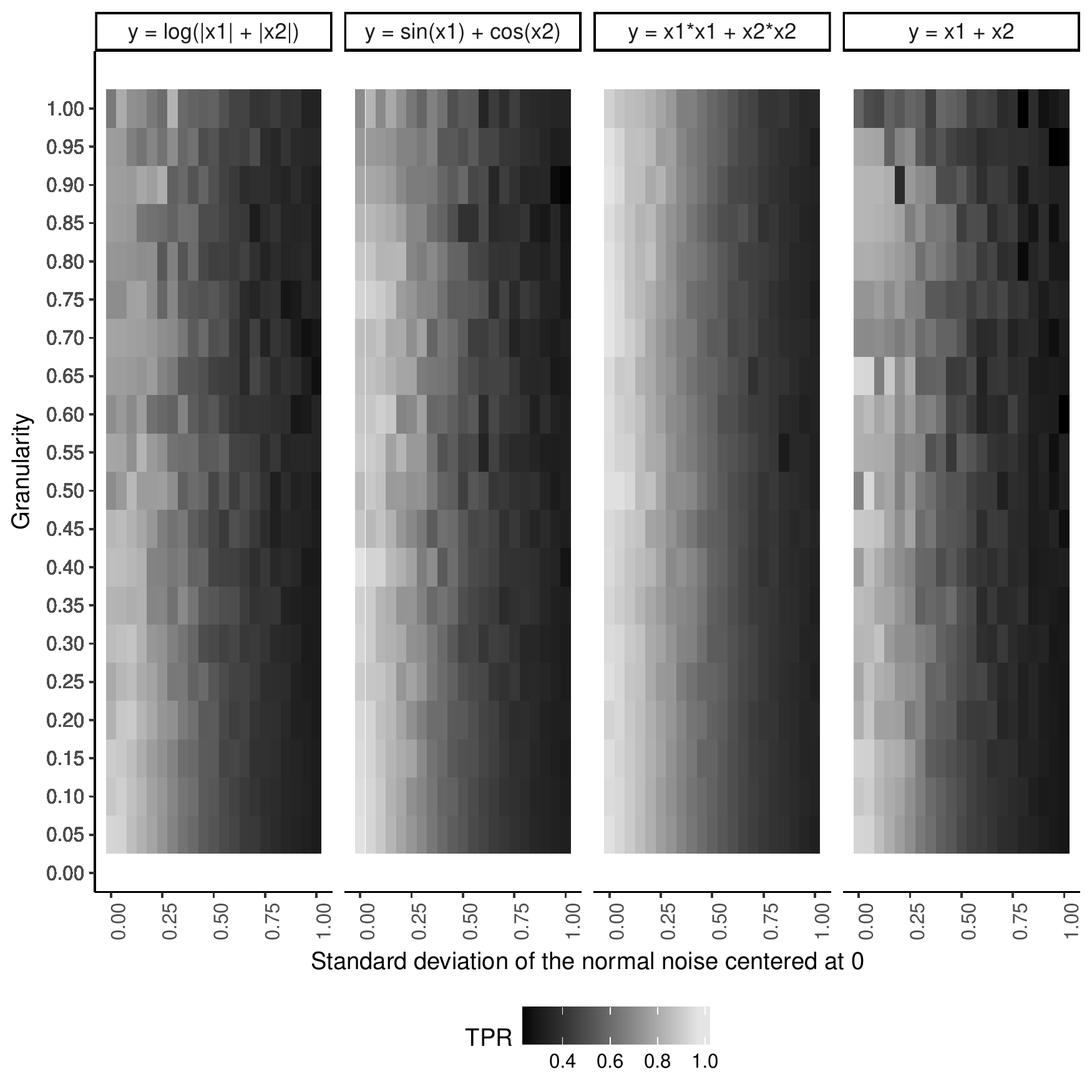}
\caption{Impact of the noise in the output on the true positive rate (TPR) of the approximate positive domains by granularity}
\label{fig:noisyoutputTPR}
\end{figure}

In this experiment, normal noise centered at zero and with different standard deviations has been added to the output of the function. Figure \ref{fig:noisyoutputTPR} shows the mean TPR of the approximate positive domains for different values of the granularity parameter by function. It can be observed that in all cases the degree of noise in the output has an important impact on the TPR of the approximate positive domains.

\begin{figure}[H]
\centering
\begin{subfigure}{.5\textwidth}
  \centering
  \includegraphics[width=\linewidth]{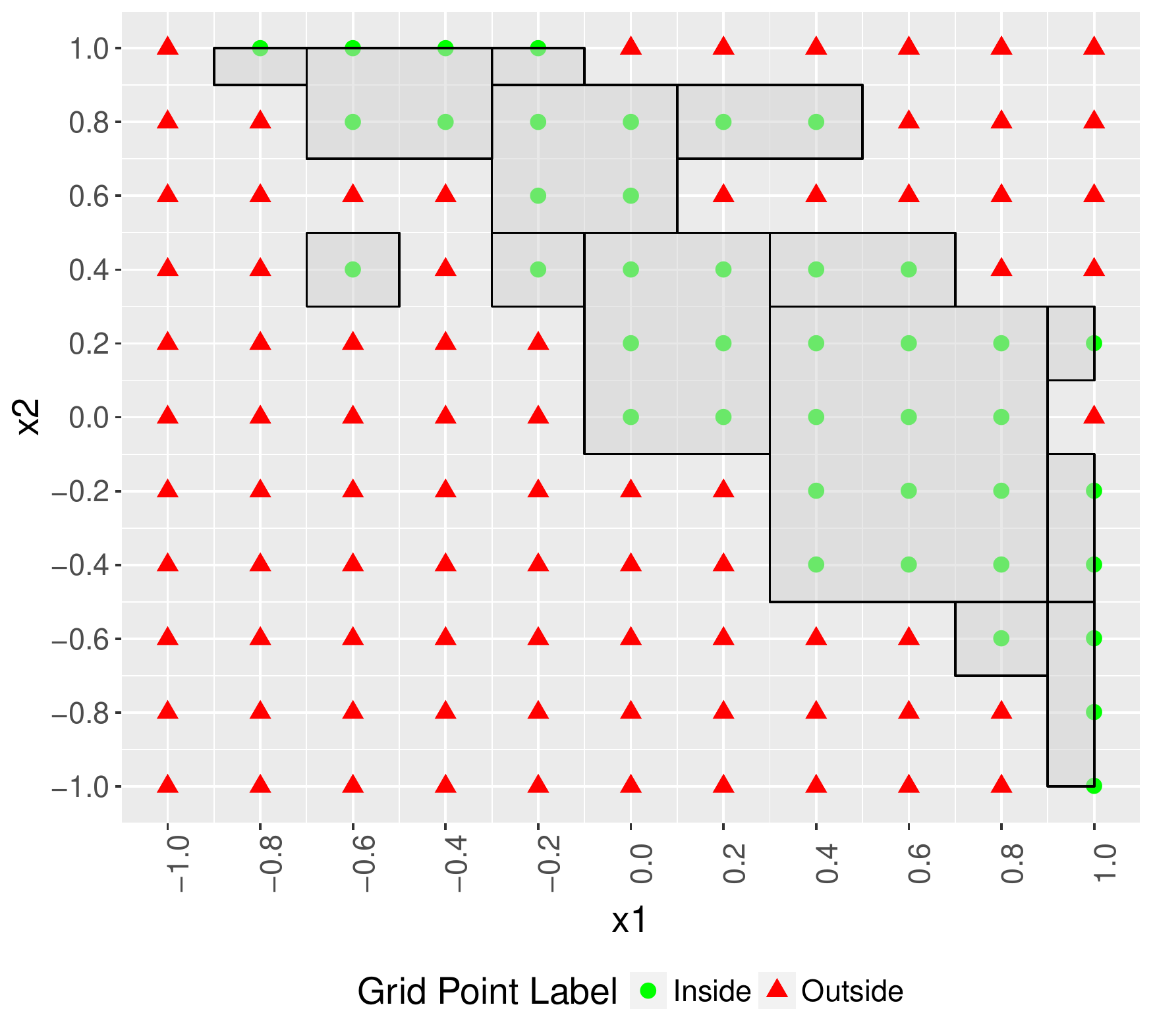}
  \caption{$y = x_1+x_2$}
  \label{fig:noisyoutput_linear_noisygrid}
\end{subfigure}%
\begin{subfigure}{.5\textwidth}
  \centering
  \includegraphics[width=\linewidth]{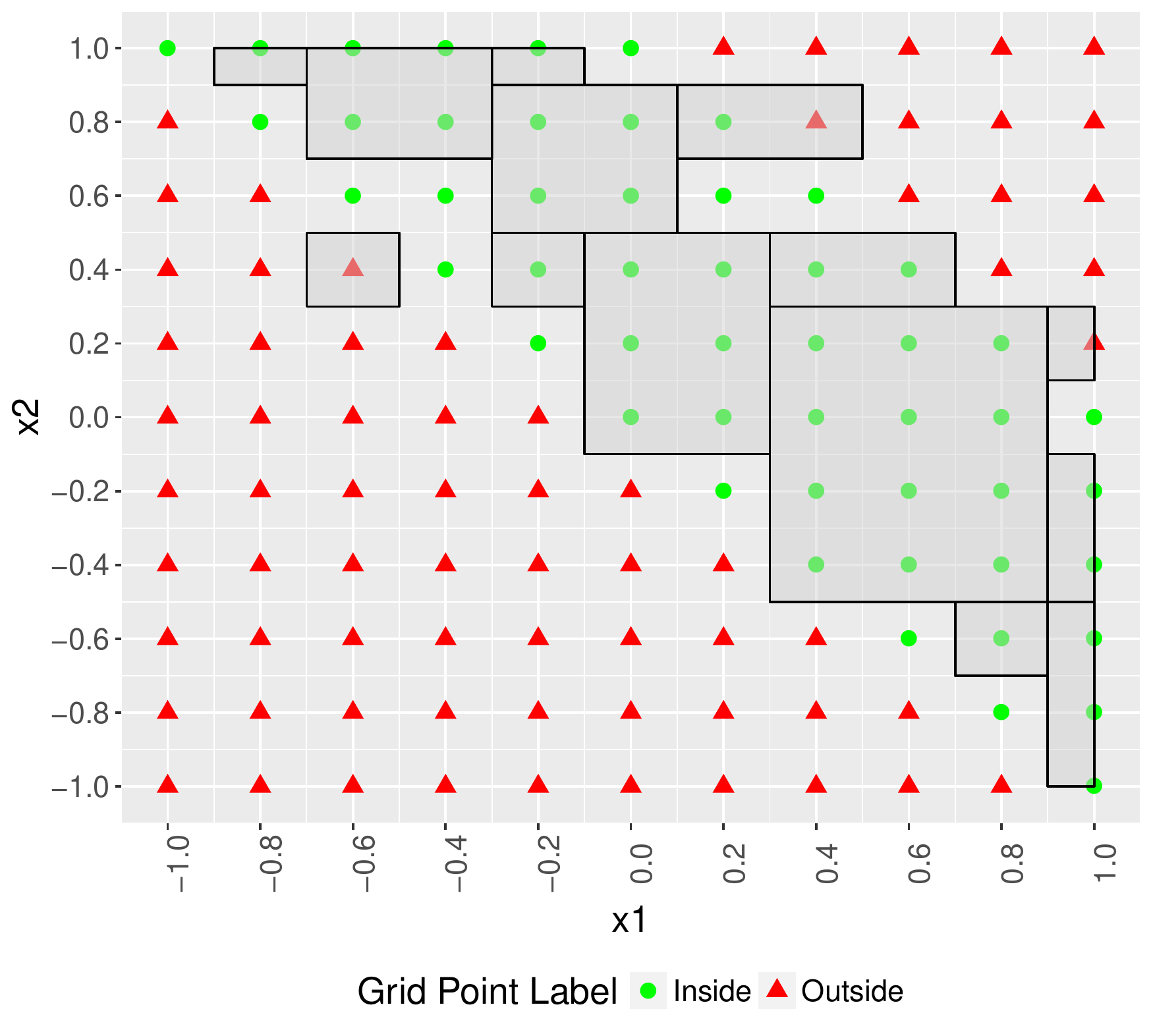}
  \caption{$y = x_1+x_2$}
  \label{fig:noisyoutput_linear_actualgrid}
\end{subfigure}
\begin{subfigure}{.5\textwidth}
  \centering
  \includegraphics[width=\linewidth]{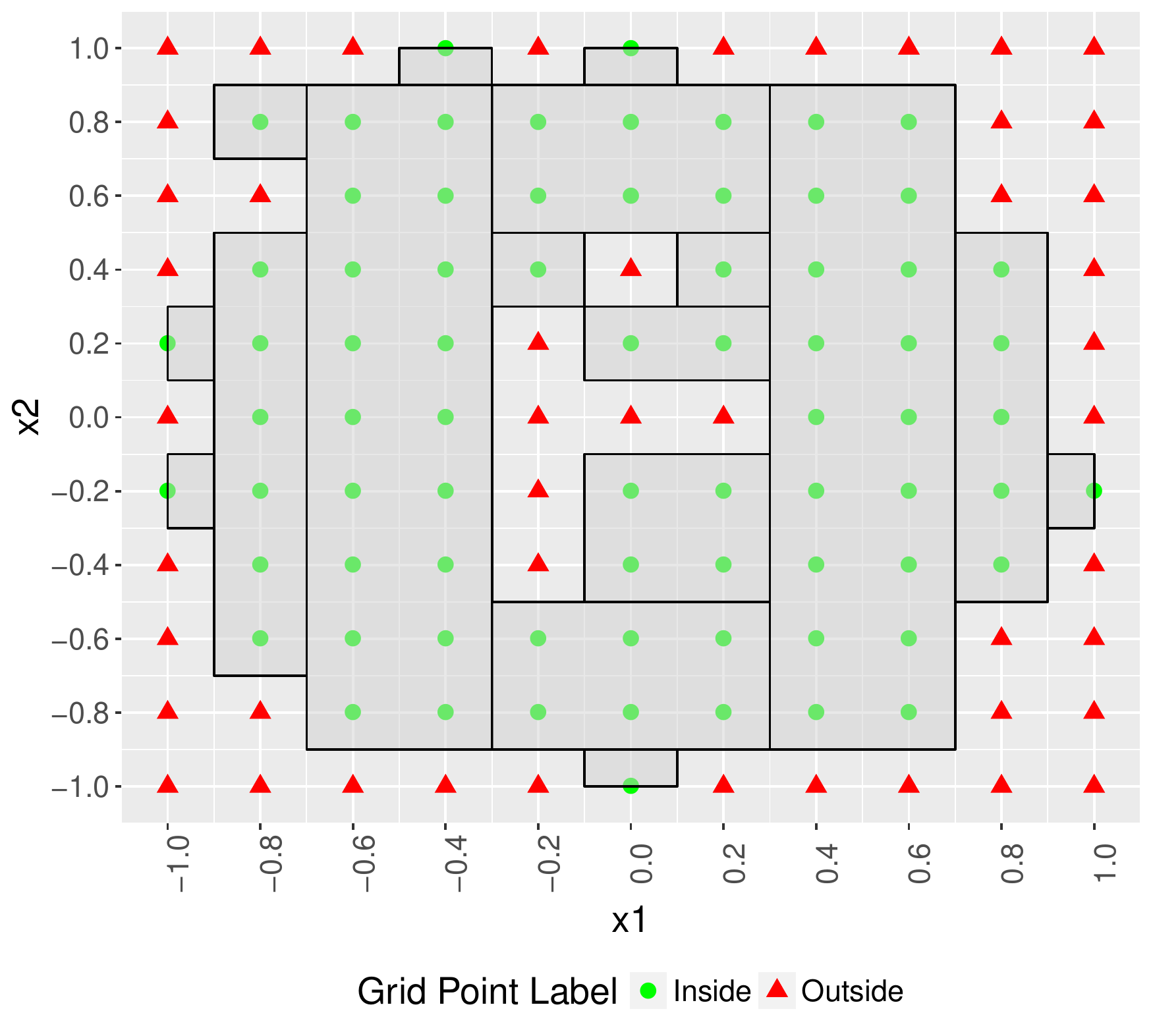}
  \caption{$y = x_1^2+x_2^2$}
  \label{fig:noisyoutput_sumofsquares_noisygrid}
\end{subfigure}%
\begin{subfigure}{.5\textwidth}
  \centering
  \includegraphics[width=\linewidth]{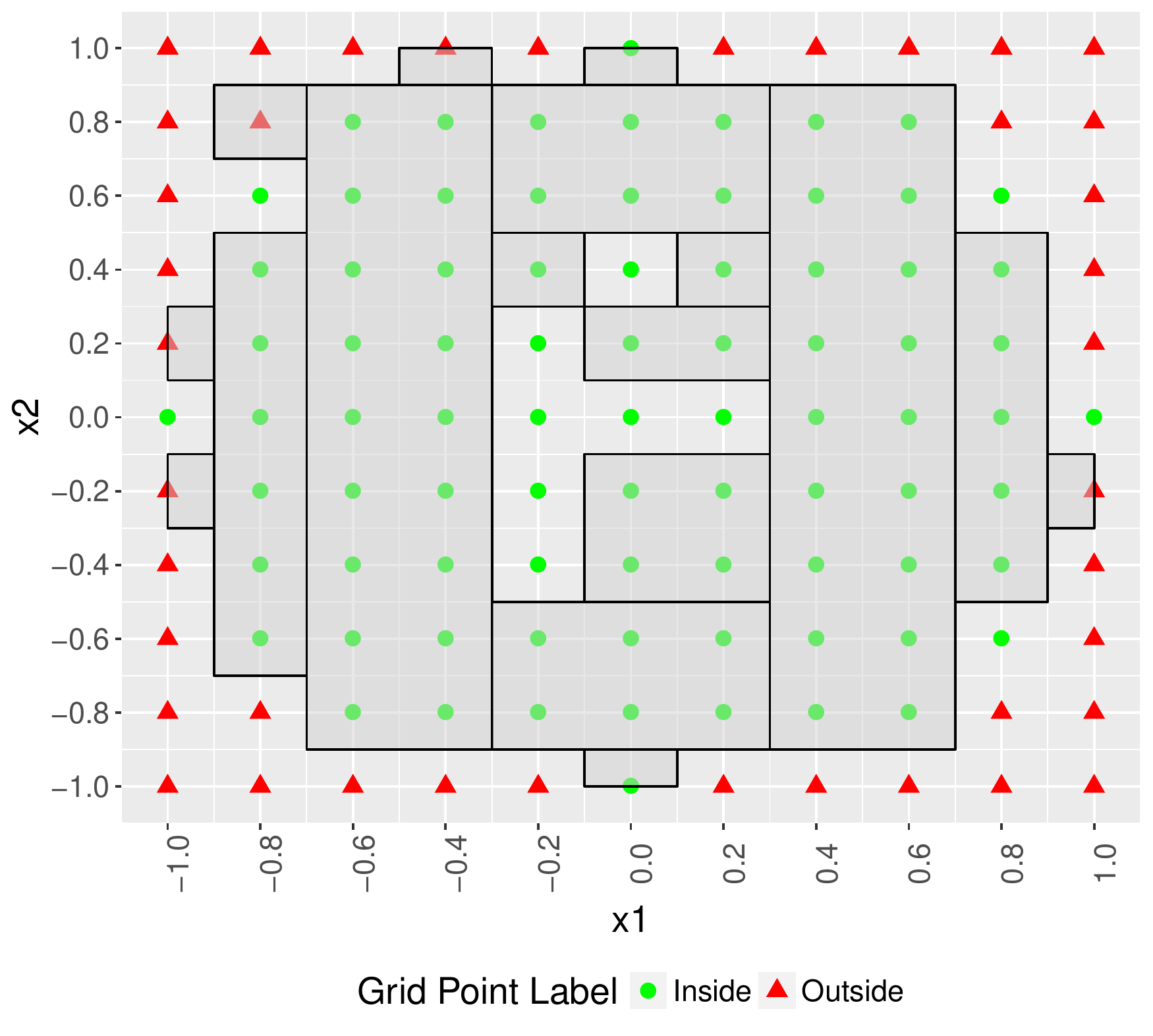}
  \caption{$y = x_1^2+x_2^2$}
  \label{fig:noisyoutput_sumofsquares_actualgrid}
\end{subfigure}
\caption{Examples of approximate positive domains with noisy outputs (granularity $\delta = 0.2$ and standard deviation of normal noise set to 0.2). Left: synthetic grid and approximate positive domain. Right: the approximate positive domain overlaid on the synthetic grid with the same granularity but no noise.}
\label{fig:noiseoutputexplanation}
\end{figure}

Figure \ref{fig:noiseoutputexplanation} illustrates what is happening with two examples. The figures on the left show the synthetic training data set created with the noisy function/model and the approximate positive domains for the $y=x_1 + x_2$ function (\ref{fig:noisyoutput_linear_noisygrid}) and for the $y= x_1^2 + x_2^2$ function (\ref{fig:noisyoutput_linear_noisygrid}). The figures on the right show the same approximate positive domains as on the corresponding figure on the left but overlaid on top of the synthetic training data set constructed using the same value for the granularity but with no noise (fig. \ref{fig:noisyoutput_linear_actualgrid} for the $y=x_1 + x_2$ function and  fig. \ref{fig:noisyoutput_linear_actualgrid} for the $y= x_1^2 + x_2^2$ function). It can be observed that, as the proposed method uses the function/model with its noisy output to construct the synthetic training data set and approximate the positive domain, both its accuracy and the TPR are negatively affected.

\subsubsection{Noisy inputs} 

The objective of this second experiment about noise is to analyse the case in which the selected values for the independent variables might not be the values that are actually used in the process (e.g. due to controller/actuator imprecision). Note that, in this case, as the output is not noisy the constructed synthetic training data set and the approximate positive domain are not affected by the noise in the inputs. In this experiment, again, the performance of the approximate positive domain is evaluated on five test sets and the mean TPR is used as the metric. The construction of these test sets is as follows. The independent variables are sampled following a uniform distribution in each variable's initial input range as before. These represent the values specified by, for example, an operator. Next, for each independent variable a corresponding noisy variable is created by adding normal noise centered at zero to the sampled variable values. Different standard deviation values for the normal noise are used to analyse how the performance changes according to the magnitude of the noise. The noisy variables represent the actual values that are in use in the process and, thus, the function output is calculated using the noisy inputs. The confusion matrix is constructed using the noiseless input values, as these were the intended values, and the function output calculated with the noisy inputs.

\begin{figure}[H]
\centering
\includegraphics[width=\linewidth]{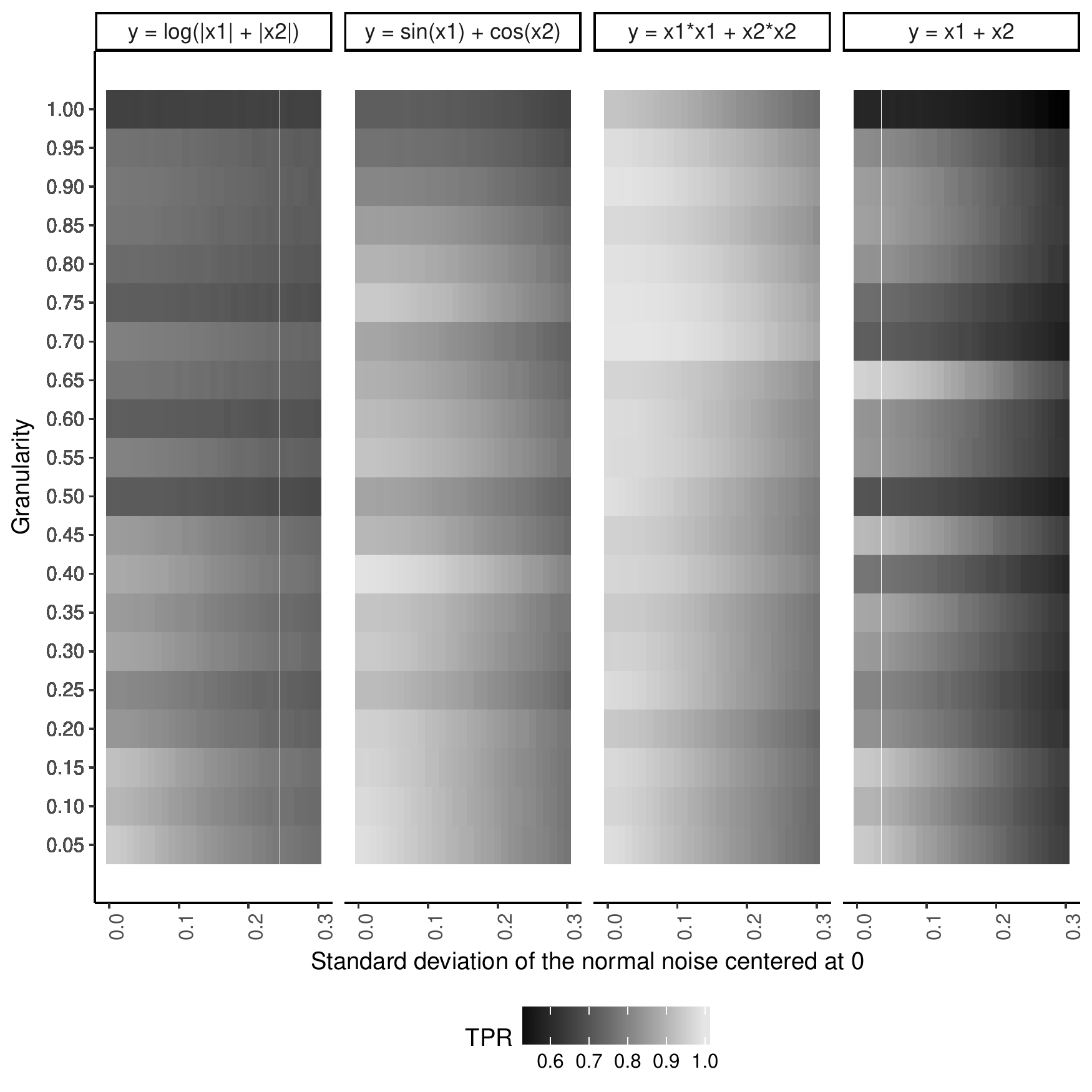}
\caption{True positive rate (TPR) of the approximate positive domains on test data with noisy inputs by granularity}
\label{fig:noisyinputsTPR}
\end{figure}

Figure \ref{fig:noisyinputsTPR} shows the mean TPR of the approximate positive domains for different levels of noise and different values of the granularity parameter by function. It shows that the noise in the inputs has a greater impact on the performance for some functions than for others. This will be analysed through some examples later.

\begin{figure}[H]
\centering
\includegraphics[width=\linewidth]{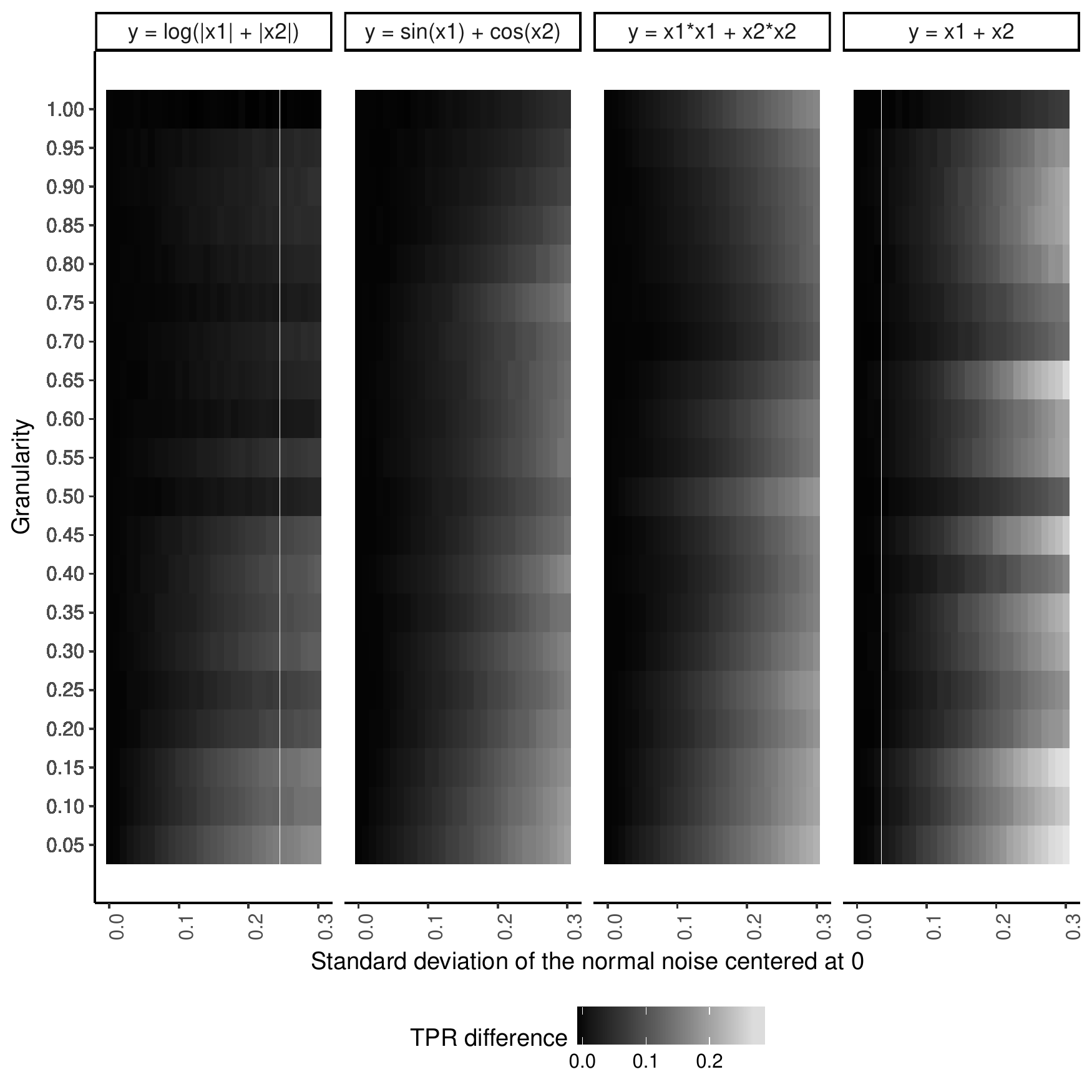}
\caption{Difference between the approximate positive domain's TPR for noiseless inputs and the approximate positive domain's TPR for noisy inputs by granularity value and function}
\label{fig:noisyinputsDIFF}
\end{figure}

Figure \ref{fig:noisyinputsDIFF} shows the difference between the approximate positive domain's TPR for noiseless inputs and the approximate positive domain's TPR for noisy inputs by granularity value and function. It can be observed that as the noise level increases for each granularity value the TPR tends to decrease, except for those granularity values for which the TPR was bad even when there was no noise (compare against fig. \ref{fig:noisyinputsTPR}).

\begin{figure}[H]
\centering
\begin{subfigure}{.35\textwidth}
  \centering
  \includegraphics[width=\textwidth]{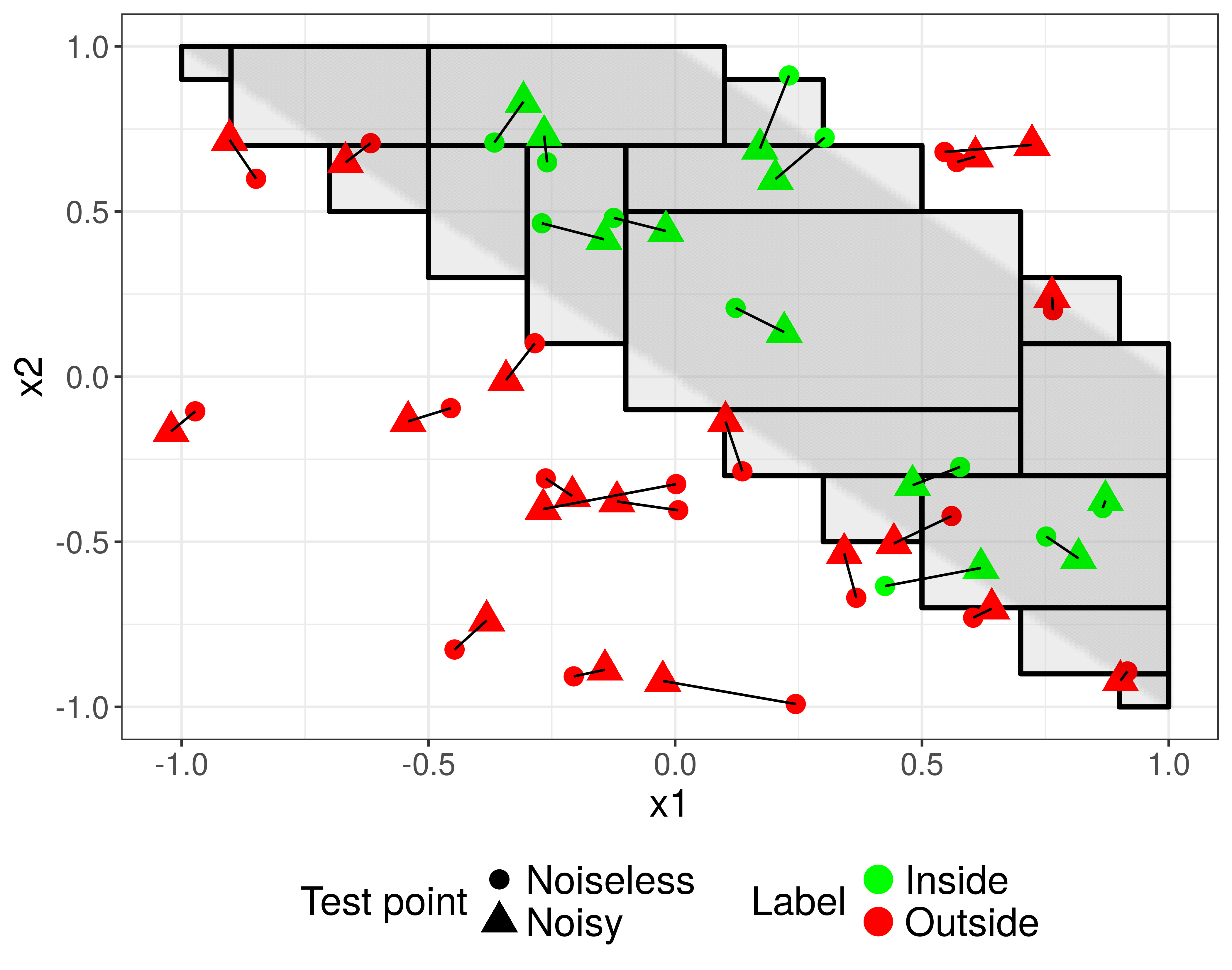}
  \caption{Standard deviation of noise: 0.1}
  \label{fig:noiseinputsexplanationlinearnoise005}
\end{subfigure}%
\begin{subfigure}{.35\textwidth}
  \centering
  \includegraphics[width=\textwidth]{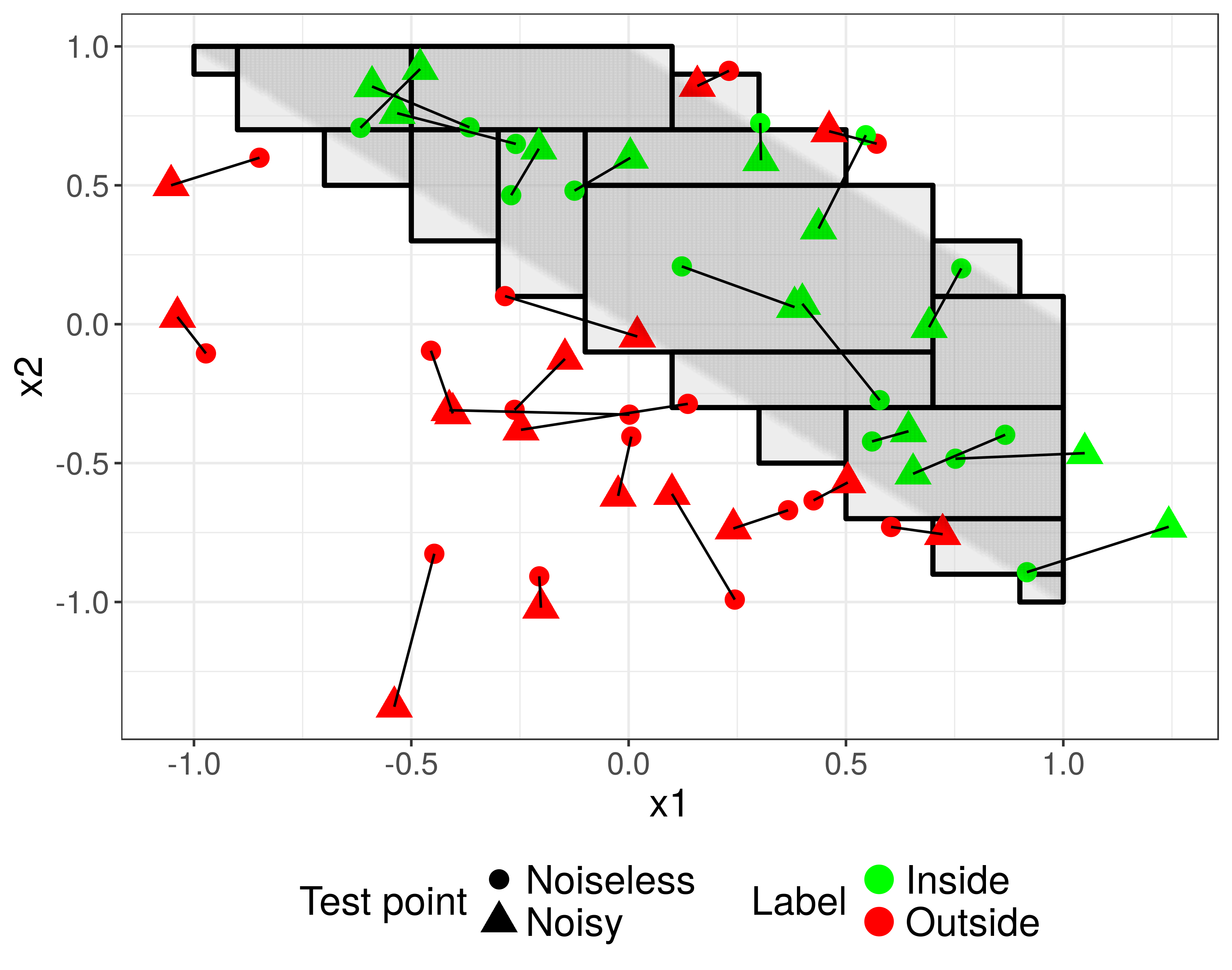}
  \caption{Standard deviation of noise: 0.2}
  \label{fig:noiseinputsexplanationlinearnoise01}
\end{subfigure}
\begin{subfigure}{.35\textwidth}
  \centering
  \includegraphics[width=\textwidth]{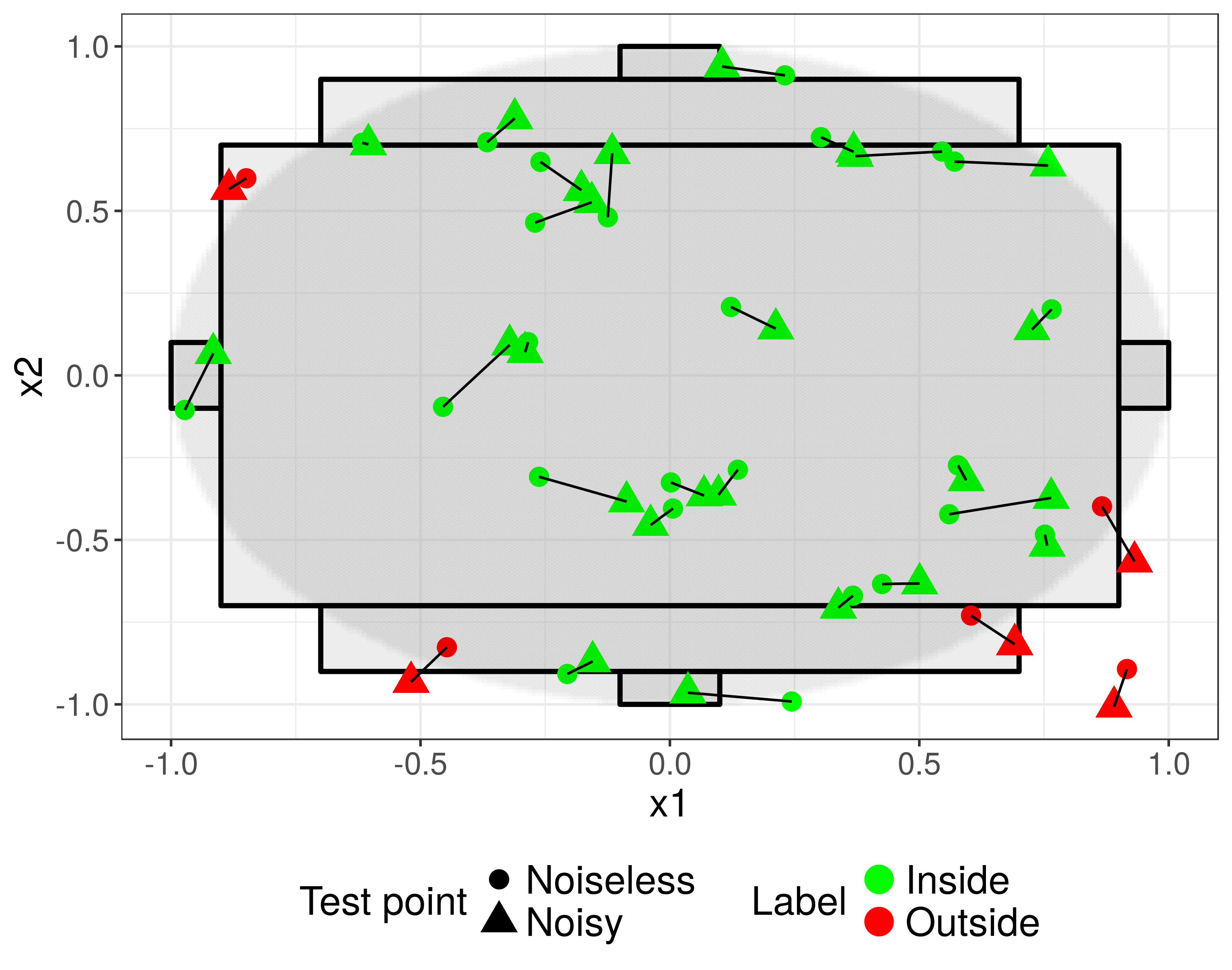}
  \caption{Standard deviation of noise: 0.1}
  \label{fig:noiseinputsexplanationsumofsquaresnoise005}
\end{subfigure}%
\begin{subfigure}{.35\textwidth}
  \centering
  \includegraphics[width=\textwidth]{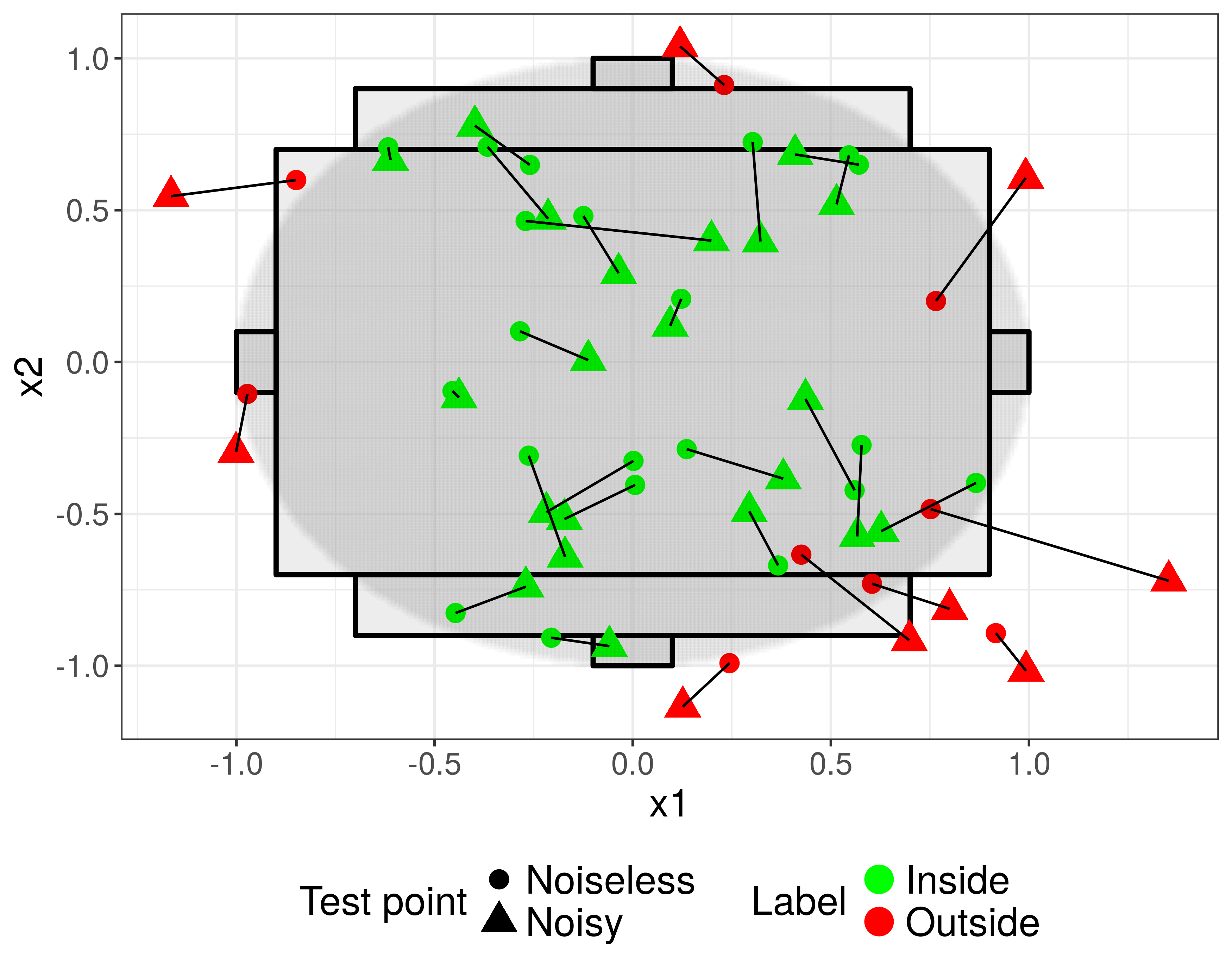}
  \caption{Standard deviation of noise: 0.2}
  \label{fig:noiseinputsexplanationsumofsquaresnoise01}
\end{subfigure}
\begin{subfigure}{.35\textwidth}
  \centering
  \includegraphics[width=\textwidth]{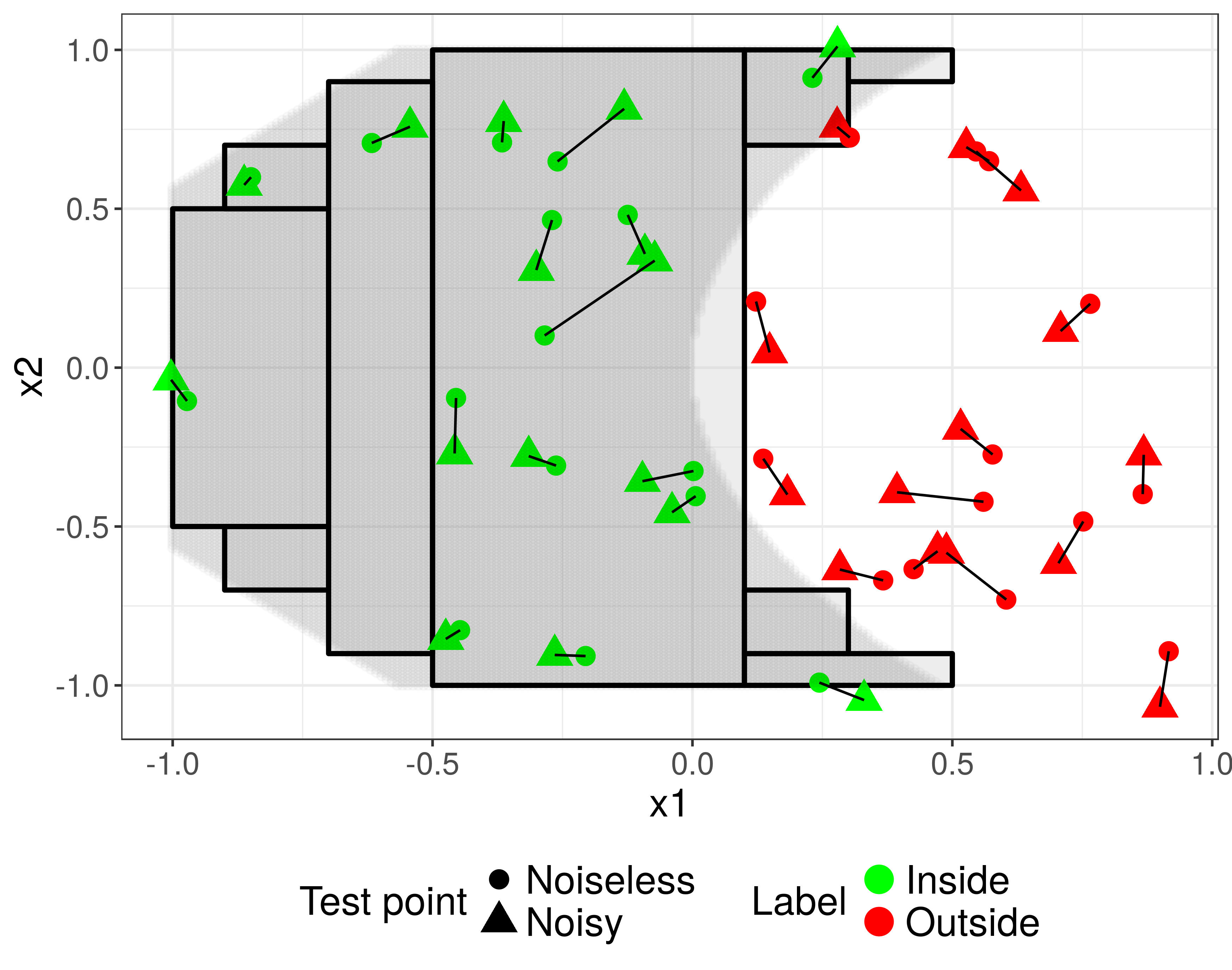}
  \caption{Standard deviation of noise: 0.1}
  \label{fig:noiseinputsexplanationsinpluscosnoise005}
\end{subfigure}%
\begin{subfigure}{.35\textwidth}
  \centering
  \includegraphics[width=\textwidth]{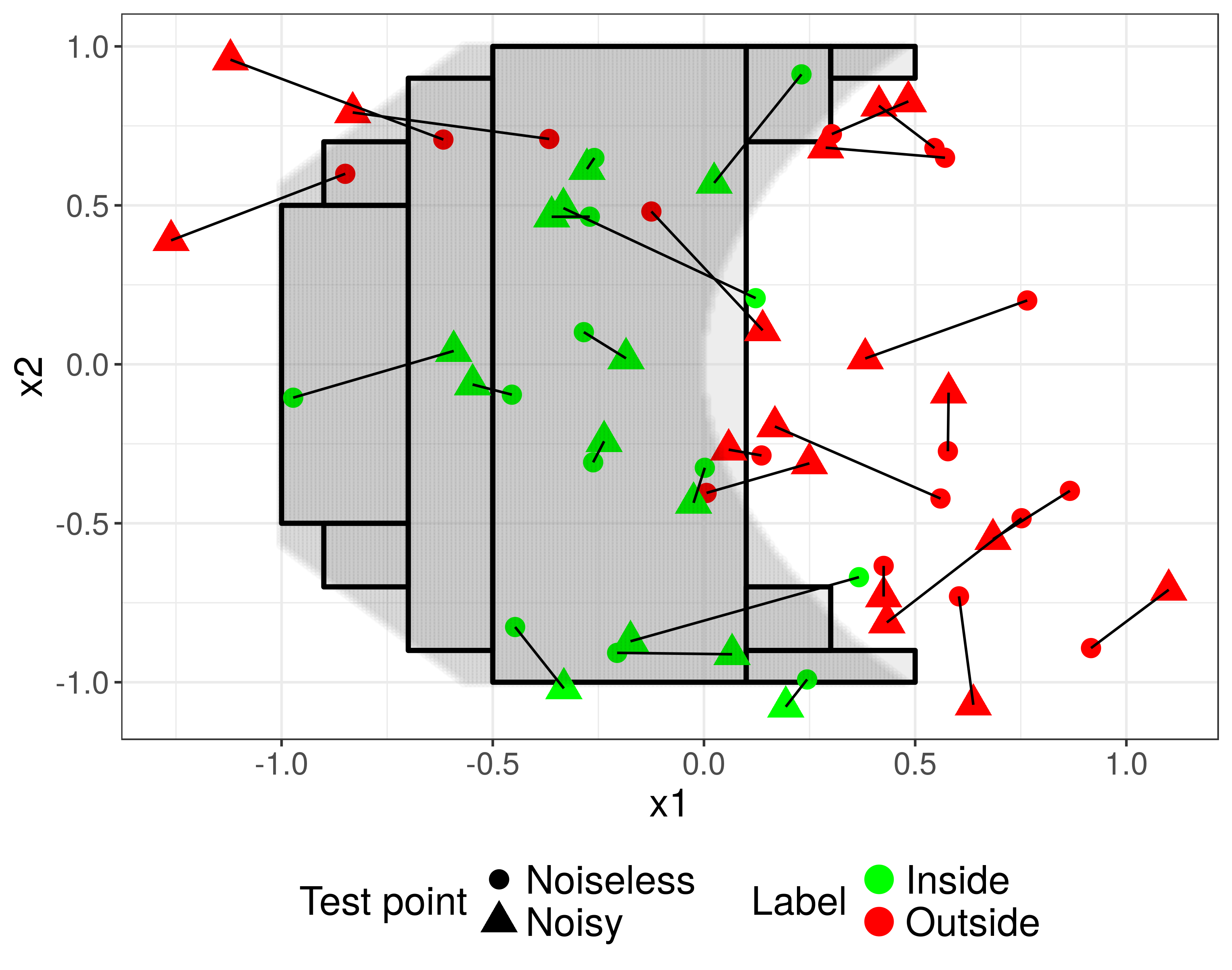}
  \caption{Standard deviation of noise: 0.2}
  \label{fig:noiseinputsexplanationsinpluscosnoise01}
\end{subfigure}
\begin{subfigure}{.35\textwidth}
  \centering
  \includegraphics[width=\textwidth]{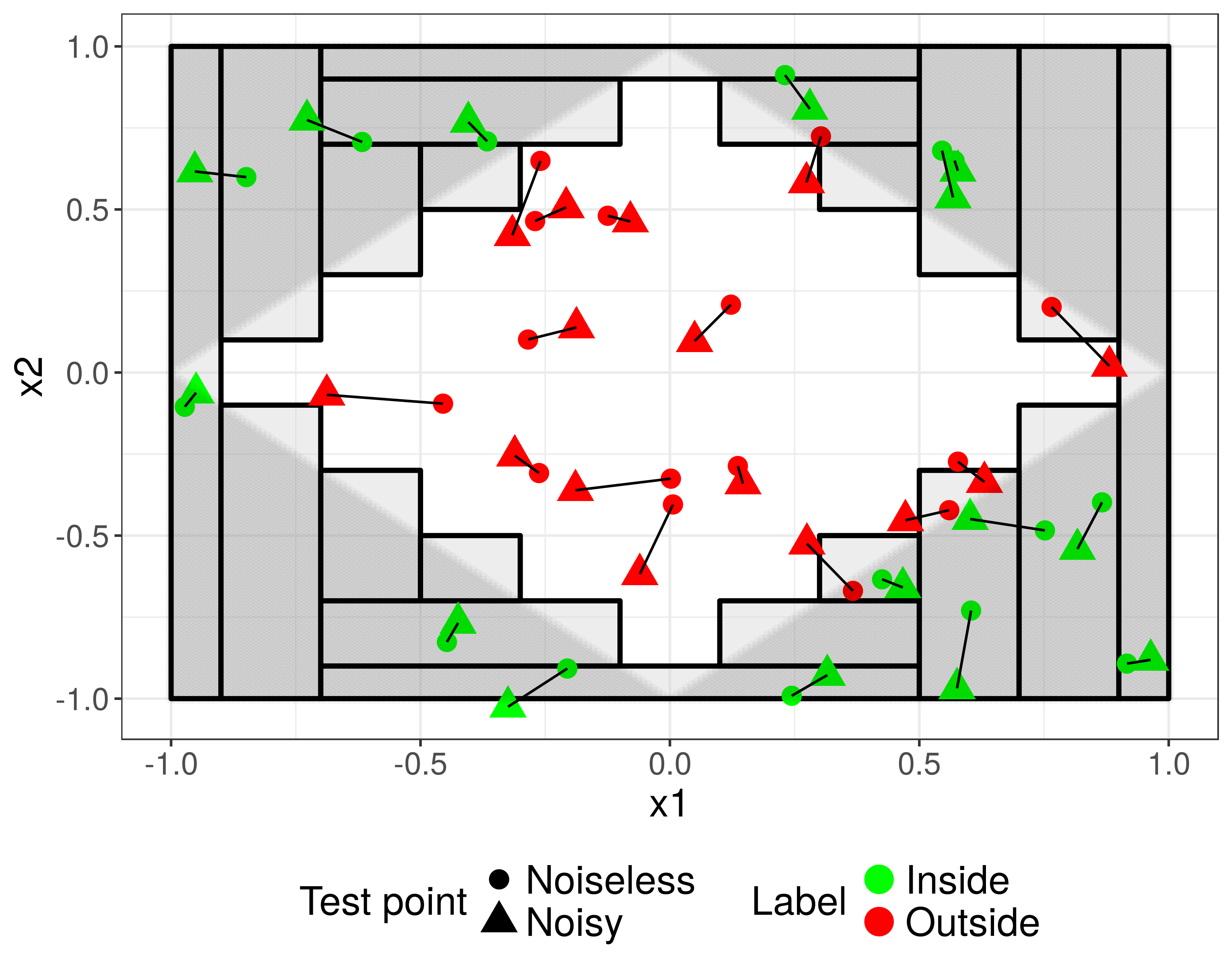}
  \caption{Standard deviation of noise: 0.1}
  \label{fig:noiseinputsexplanationlogsumabsnoise005}
\end{subfigure}%
\begin{subfigure}{.35\textwidth}
  \centering
  \includegraphics[width=\textwidth]{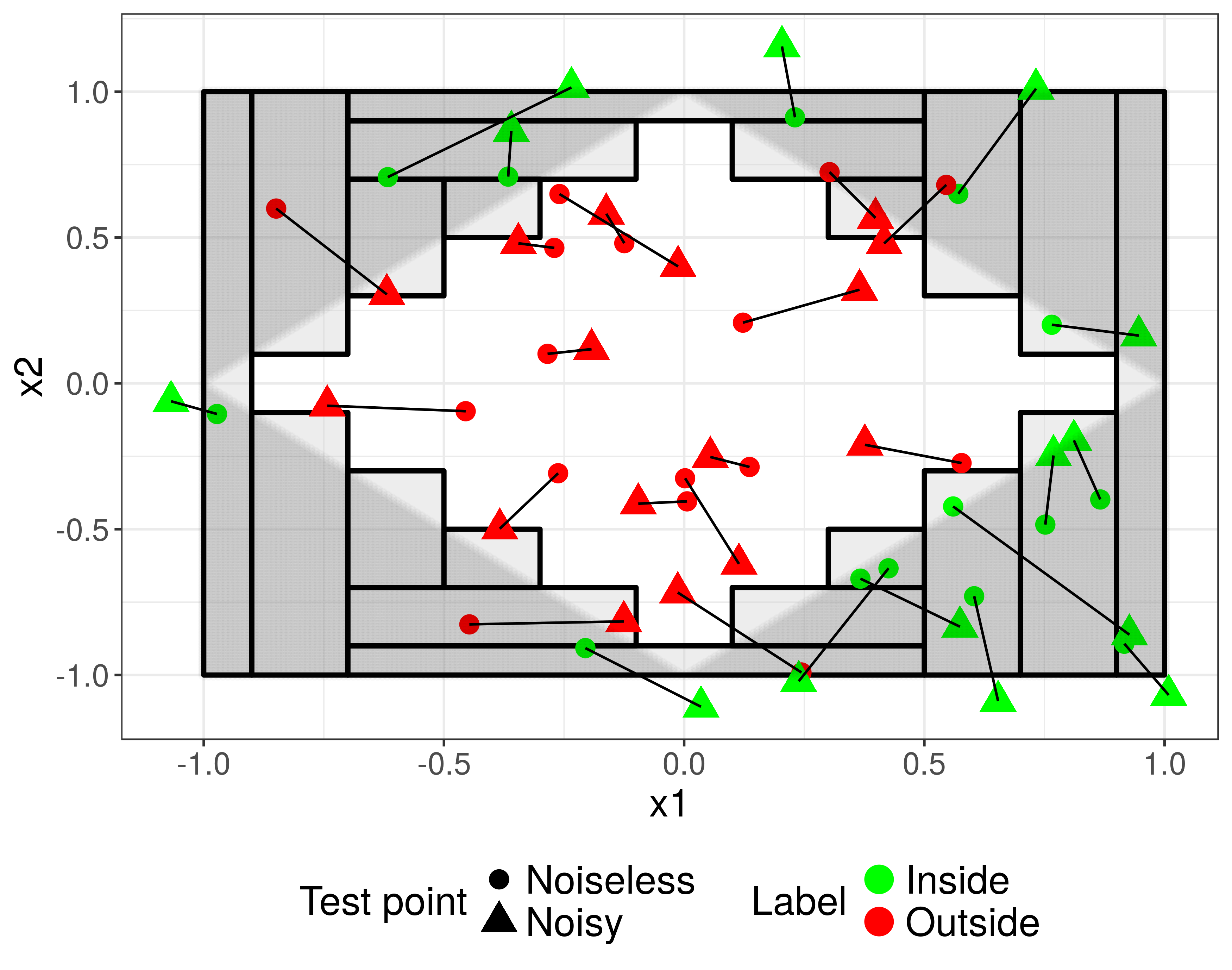}
  \caption{Standard deviation of noise: 0.2}
  \label{fig:noiseinputsexplanationlogsumabsnoise01}
\end{subfigure}
\caption{Effect of noisy inputs. Function: first row $y = x_1 + x_2$, second row $y = x_1^2 + x_2^2$, third row $y = \sin(x_1) + \cos(x_2)$, fourth row $y = \log( \lvert x_1 \lvert + \lvert x_2 \lvert )$. Granularity: $\delta = 0.2$, in all cases. Shaded area: positive domain. Rectangles: approximate positive domain.}
\label{fig:noiseinputsexplanation}
\end{figure}

Figure \ref{fig:noiseinputsexplanation} illustrates the effect of noisy inputs. The first row (figs. \ref{fig:noiseinputsexplanationlinearnoise005} and \ref{fig:noiseinputsexplanationlinearnoise01}) shows two examples with different standard deviation for the noise for the $f = x_1 + x_2$ function. Similarly, the second row (figs. \ref{fig:noiseinputsexplanationsumofsquaresnoise005} and \ref{fig:noiseinputsexplanationsumofsquaresnoise01}) shows two examples for the $f = x_1^2 + x_2^2$ function. The third row (figs. \ref{fig:noiseinputsexplanationsinpluscosnoise005} and \ref{fig:noiseinputsexplanationsinpluscosnoise01}) shows examples for $f = \sin(x_1) + \cos(x_2)$. Lastly, figs. \ref{fig:noiseinputsexplanationsinpluscosnoise005} and \ref{fig:noiseinputsexplanationsinpluscosnoise01} on the fourth row show examples for the $f = \log( \lvert x_1 \lvert + \lvert x_2 \lvert )$. The shaded areas represent the positive domain and the rectangles represent the approximate positive domain in each case. The noisy inputs are represented linked to the noiseless input they correspond to. The label of the two points is determined by whether the function output for the noisy input is within the target output range or not. 
Let's define as the \emph{initial set} the cartesian product of initial input ranges for the independent variables. The initial set can be divided into four disjoint subsets. The \emph{true positive set} is the intersection between the positive domain and the approximate positive domain. The \emph{true negative set} is the intersection between the negative domain and the complement of the approximate positive domain relative to the initial set. The \emph{false positive set} is the intersection between the negative domain and the approximate positive domain. Lastly, the \emph{false negative set} is the intersection between the positive domain and the complement of the approximate positive domain. It is possible that one or more of these subsets may be the empty set. In figs. \ref{fig:noiseinputsexplanationlinearnoise005} and \ref{fig:noiseinputsexplanationlinearnoise01}, for example, the false negative set is empty. When a noisy point and its corresponding noiseless point are both within the same subset of the initial set, the noise has not affected the label of the point. However, when the noisy and corresponding noiseless points are within different subsets of the initial set, they have different labels and, thus, the noise in the inputs has had an effect on the label of the output. Figure \ref{fig:noiseinputsexplanation} provides different examples of the different situations that can occur. Given a noiseless point inside the true positive set, whether the corresponding noisy point is outside the true positive set depends on the magnitude of the noise but, also, on the size, geometry, and topology of the true positive set. The latter factors are responsible for the noise affecting the TPR differently for different functions.

%%%%%%%%%%%%%%%%%%%%%%%%%%%%%%%%%%%%%%%%%%%%%%%%%%%%%%%%%%%%%%%%%%%%%%%%%%%%%%%%%%%%%%%%%%%%%%%%%%%%%%%%%%
%%%%%%%%%%%%%%%%%%%%%%%%%%%%%%%%%%%%%%%%%%%%%%%%%%%%%%%%%%%%%%%%%%%%%%%%%%%%%%%%%%%%%%%%%%%%%%%%%%%%%%%%%%
\section{Results with real data}
\label{sec:results}

The proposed heuristic has been validated using real data from a manufacturing process to produce parts. The client needs to keep the eccentricity of the produced parts below a threshold. The process is controlled with four actuators. Using real data of this process a regression model has been trained that estimates the output variable, i.e. the eccentricity, based on the values of the input variables, i.e. the actuators. The initial ranges for the input variables are $[0.0, 6.0]$ for $x_1$, $[0.0, 5.0]$ for $x_2$, $[0.0, 2.0]$ for $x_3$, and $[1.0, 5.0]$ for $x_4$. The client requires the eccentricity to be below 10. Therefore, the desired output range is $[0,10)$. However, because the trained model has an out-of-sample root mean square error (RMSE) of 0.97, the target output range passed to the proposed method is $[0,9)$, to account for the model error.

Tables \ref{tab:realresultsbefore} and \ref{tab:realresultsafter} show the confusion matrix for the approximate positive domain before and after, respectively, removing parts of the approximate positive domain to improve the TPR as described in section \ref{sec:improvingtpr}. The TPR of the original approximate positive domain is 0.9875 and the improved TPR is 0.98992. The test set has 27 defective parts (i.e. parts with eccentricity greater than or equal to 10) out of a total of 1146 parts, thus 2.36 \% of the parts in the test set are defective. Taking only those parts in the test set for which the input values are within the approximate positive domain, there are only 14 defective parts out of a total of 1120 parts, i.e. 1.25 \% of the parts in this subset of the test set. Finally, taking only those parts in the test set for which their input values are within the improved TPR approximate positive domain, then there are 11 defective parts out of a total of 1091 parts, 1.01 \% of the parts in this second subset of the test set. So there has been an improvement of 1.35 \%, from 2.36 \% to 1.01 \%.

\begin{table}[H]
\centering
\caption{Evaluation of the approximate positive domain on the test set before removing parts of the approximate positive domain to improve TPR}
\label{tab:realresultsbefore}
\begin{tabular}{c|c|c|c|}
\multicolumn{1}{c}{} & \multicolumn{1}{c}{} & \multicolumn{2}{c}{OUTPUT} \\ 
\cline{3-4}
\multicolumn{1}{c}{} & & Inside & Outside  \\
\cline{2-4}
\multirow{2}{*}{INPUT} & Inside & 1106 & 14 \\
\cline{2-4}
& Outside & 13 & 13 \\
\cline{2-4}
\end{tabular}
\end{table}

\begin{table}[H]
\centering
\caption{Evaluation of the approximate positive domain on the test set after removing parts of the approximate positive domain to improve TPR}
\label{tab:realresultsafter}
\begin{tabular}{c|c|c|c|}
\multicolumn{1}{c}{} & \multicolumn{1}{c}{} & \multicolumn{2}{c}{OUTPUT} \\ 
\cline{3-4}
\multicolumn{1}{c}{} & & Inside & Outside  \\
\cline{2-4}
\multirow{2}{*}{INPUT} & Inside & 1080 & 11 \\
\cline{2-4}
& Outside & 39 & 16 \\
\cline{2-4}
\end{tabular}
\end{table}

\begin{figure}[H]
\centering
\begin{subfigure}{.495\textwidth}
  \centering
  \includegraphics[width=\linewidth]{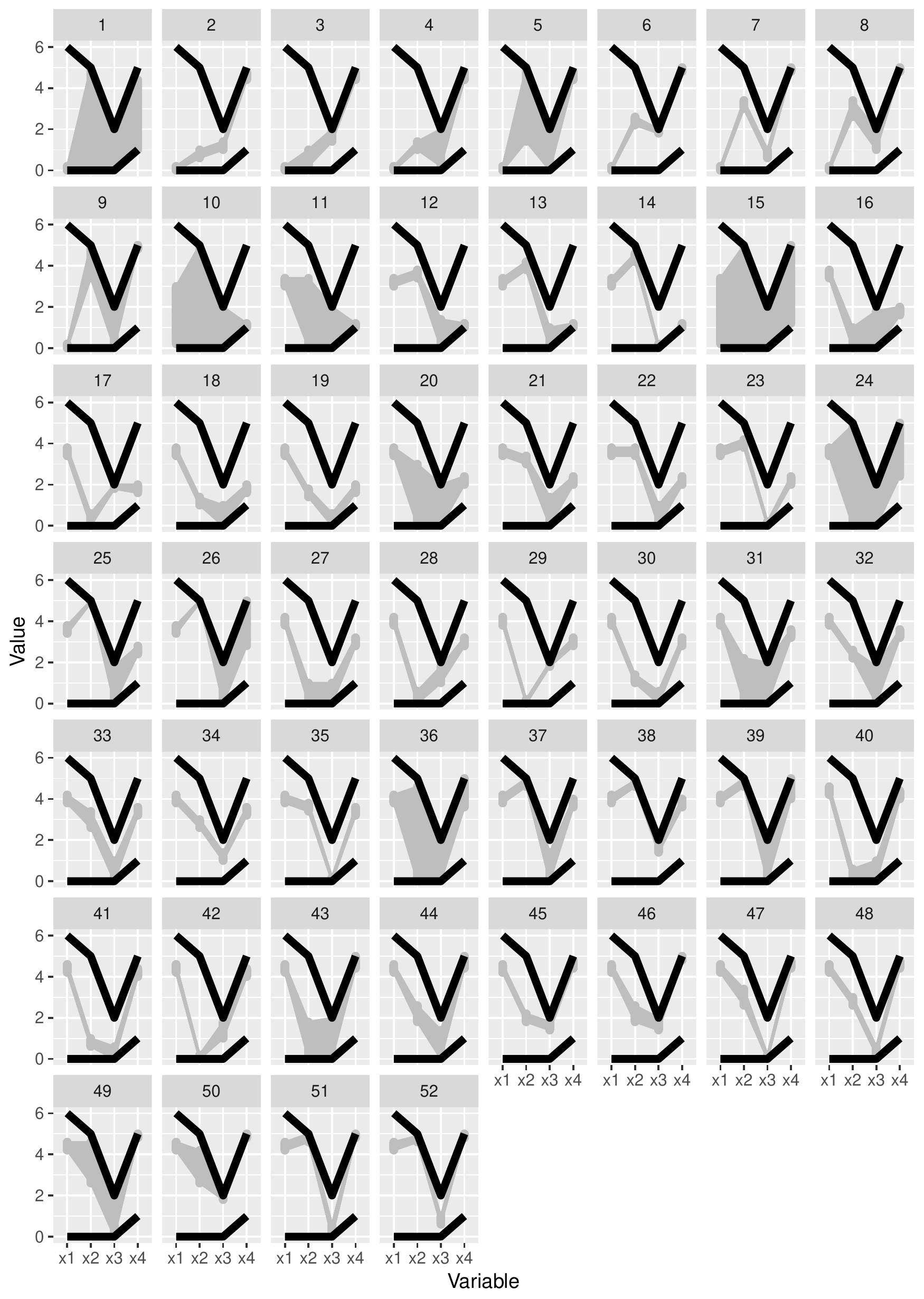}
  \caption{Original}
  \label{fig:realdataintervals1}
\end{subfigure}
\begin{subfigure}{.495\textwidth}
  \centering
  \includegraphics[width=\linewidth]{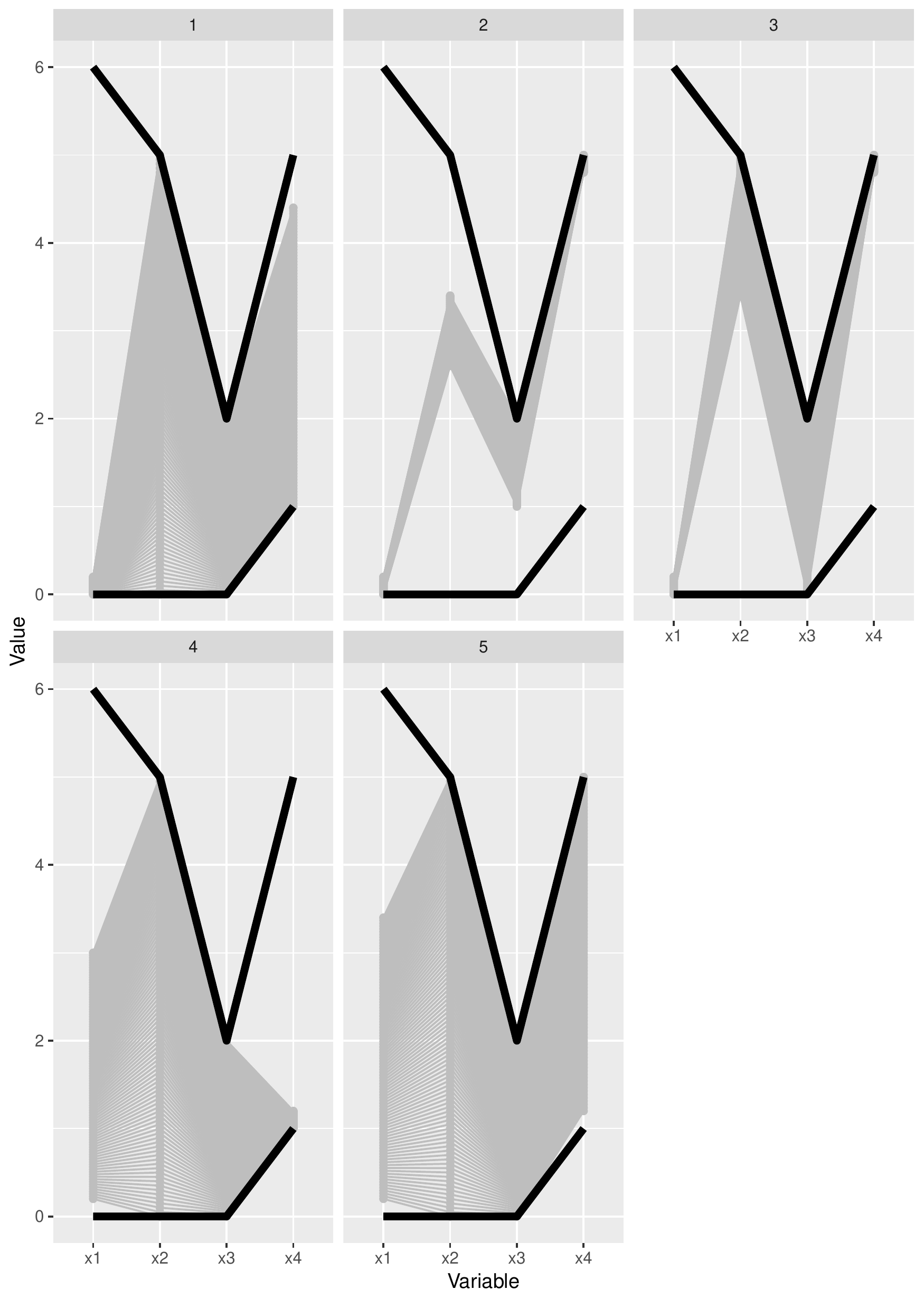}
  \caption{Improved TPR}
  \label{fig:realdataintervals2}
\end{subfigure}
\caption{Approximate positive domain}
\label{fig:realdataintervals}
\end{figure}

Figure \ref{fig:realdataintervals} shows the obtained approximate positive domain as the grey shaded area in between the black lines, which represent the boundaries of the initial ranges for the inputs. Fig. \ref{fig:realdataintervals1} shows the original approximate positive domain with the granularity parameter set to 0.4 and fig. \ref{fig:realdataintervals2} shows the improved TPR approximate positive domain.

%%%%%%%%%%%%%%%%%%%%%%%%%%%%%%%%%%%%%%%%%%%%%%%%%%%%%%%%%%%%%%%%%%%%%%%%%%%%%%%%%%%%%%%%%%%%%%%%%%%%%%%%%%
%%%%%%%%%%%%%%%%%%%%%%%%%%%%%%%%%%%%%%%%%%%%%%%%%%%%%%%%%%%%%%%%%%%%%%%%%%%%%%%%%%%%%%%%%%%%%%%%%%%%%%%%%%
\section{Conclusions and future work}
\label{sec:conclusions}

The proposed heuristic provides a solution to a problem present in different areas of manufacturing: determining input variable ranges for a given output range. The method proposed creates a synthetic, labeled training data set of input combinations in which the class label indicates whether the output produced by the function applied to each input combination is within the given target range or not. This is a classification problem and a decision tree classifier is trained with the newly created training set. The trained decision tree classifier provides combinations of input ranges labeled depending on whether the output will be inside or outside the given range. The union of the products of input ranges for which the output is within the given target range is an approximation to the desired set. The method also proposes a performance indicator to evaluate how good the obtained set is, by estimating the ratio of points in the set for which the value the function takes at the point is within the target output range. The presented method assumes the target output range is a closed interval but can be trivially extended to any kind of interval, even intervals with only one finite endpoint, or set. Additionally, the solution applies to both mathematical functions, invertible and non-invertible, and empirical models, which may (e.g. linear regression) or may not (e.g. regression tree) have a mathematical expression. The solution provides good performance and its behaviour is well understood. Finally, a way to improve the performance in terms of the true positive rate has been proposed which consists in reducing the magnitude of the granularity parameter value so that the obtained set better approximates the positive domain and only use those product of input ranges that are contained in the positive domain (i.e. their true positive rate is 1).
%%%%%%%%%%%%%%%%%%%%%%%%%%%%%%%%%%%%%%%%%%%%%%%%%%%%%%%%%%%%%%%%%%%%%%%%%%%%%%%%%%%%%%%%%%%%%%%%%%%%%%%%%%
%%%%%%%%%%%%%%%%%%%%%%%%%%%%%%%%%%%%%%%%%%%%%%%%%%%%%%%%%%%%%%%%%%%%%%%%%%%%%%%%%%%%%%%%%%%%%%%%%%%%%%%%%%
\bibliographystyle{elsarticle-num} 
\bibliography{prescription}
%%%%%%%%%%%%%%%%%%%%%%%%%%%%%%%%%%%%%%%%%%%%%%%%%%%%%%%%%%%%%%%%%%%%%%%%%%%%%%%%%%%%%%%%%%%%%%%%%%%%%%%%%%
%%%%%%%%%%%%%%%%%%%%%%%%%%%%%%%%%%%%%%%%%%%%%%%%%%%%%%%%%%%%%%%%%%%%%%%%%%%%%%%%%%%%%%%%%%%%%%%%%%%%%%%%%%
\end{document}